\documentclass{article}

\usepackage{colt11e}
\usepackage[round,comma]{natbib}
\usepackage{amsmath,amssymb}
\usepackage{mathrsfs}
\usepackage{mathtools}
\mathtoolsset{showonlyrefs=true}

\newenvironment{proof*}{\noindent{\bf Proof:}}{}
\newcommand{\ignore}[1]{}

\newcommand{\dd}{\mathrm{d}}
\newcommand{\calM}{\mathcal{M}}
\newcommand{\EE}{\mathrm{E}}

\newcommand{\Real}{\mathbb{R}}

\newcommand{\fhat}{\hat{f}}
\newcommand{\gstar}{g^*}
\newcommand{\fstar}{f^*}
\newcommand{\Istar}{I_0}

\newcommand{\calB}{\mathcal{B}}
\newcommand{\calE}{\mathcal{E}}
\newcommand{\calF}{\mathcal{F}}
\newcommand{\calG}{\mathcal{G}}
\newcommand{\calH}{\mathcal{H}}

\newcommand{\calN}{\mathcal{N}}

\newcommand{\calX}{\mathcal{X}}
\newcommand{\boldK}{\boldsymbol{K}}

\newcommand{\scrE}{\mathscr{E}}

\newcommand{\Eqref}[1]{Eq.~{\eqref{#1}}}

\newcommand{\kmin}{\kappa}

\newcommand{\hnorm}[1]{\|_{\calH_{#1}}}

\newcommand{\lambdaone}{{\lambda_1^{(n)}}}
\newcommand{\lambdatwo}{{\lambda_2^{(n)}}}
\newcommand{\lambdathree}{{\lambda_3^{(n)}}}
\newcommand{\lambdatmp}{{\lambda}}

\newcommand{\boldalpha}{\boldsymbol{\alpha}}

\newcommand{\LPi}{L_2(\Pi)}

\newcommand{\totH}{\calH}

\newcommand{\sprime}{\tilde{s}}

\newtheorem{Theorem}{Theorem}

\newtheorem{Lemma}[Theorem]{Lemma}
\newtheorem{Proposition}[Theorem]{Proposition}
\newtheorem{Corollary}[Theorem]{Corollary}
\newtheorem{Assumption}{Assumption}

\usepackage[dvipdfm]{graphicx}

\allowdisplaybreaks

\title{Fast Convergence Rate of Multiple Kernel Learning \\ with Elastic-net Regularization}
\author{Taiji Suzuki, Ryota Tomioka \\
Department of Mathematical Informatics, \\
The University of Tokyo,\\
7-3-1 Hongo, Bunkyo-ku, Tokyo\\
\texttt{\small t-suzuki@mist.i.u-tokyo.ac.jp}, \\ 
\texttt{\small tomioka@mist.i.u-tokyo.ac.jp}
\And
Masashi Sugiyama \\
Department of Computer Science, \\
Tokyo Institute of Technology,\\
2-12-1 O-okayama, Meguro-ku, Tokyo\\
\texttt{\small sugi@cs.titech.ac.jp}
}

\begin{document}

\maketitle

\begin{abstract}
We investigate the learning rate of multiple kernel leaning (MKL)
with elastic-net regularization,
which consists of an $\ell_1$-regularizer for inducing the sparsity
and an $\ell_2$-regularizer for controlling the smoothness.
We focus on a sparse setting where the total number of kernels is large
but the number of non-zero components of the ground truth is relatively small,
and prove that elastic-net MKL achieves the minimax learning rate
on the $\ell_2$-mixed-norm ball.
Our bound is sharper than the convergence rates ever shown,
and has a property that the smoother the truth is,
the faster the convergence rate is.
\end{abstract}

\section{Introduction}
Learning with kernels such as support vector machines has been demonstrated to
be a promising approach,
given that kernels were chosen appropriately
 \citep{book:Schoelkopf+Smola:2002,Book:Taylor+Cristianini:2004}.
% However, since the behavior of the kernel methods relies heavily on the choice of kernels,
% the issue of kernel selection needs to be addressed properly for making the kernel methods
% has been an important problem. 
So far, various strategies have been employed for choosing appropriate kernels,
ranging from simple cross-validation 
\citep{mach:chapelle+vapnik+bousquet:2002}
%\citep{JRSS:Stone:1974}
 to
more sophisticated `kernel learning' approaches
\citep{JMLR:Ong+etal:2005,ICML:Argriou+etal:2006,NIPS:Bach:2009,NIPS:Cortes+etal:nonlinear:2009,ICML:Varma+Babu:2009}.

\emph{Multiple kernel learning} (MKL) is one of the systematic approaches
to learning kernels, which tries to find the optimal linear combination
of prefixed base-kernels by convex optimization \citep{JMLR:Lanckriet+etal:2004}.
The seminal paper by \citet{ICML:Bach+etal:2004} showed that this linear-combination
MKL formulation can be interpreted as $\ell_1$-mixed-norm regularization 
(i.e., the sum of the norms of the base kernels).
Based on this interpretation,
several variations of MKL were proposed,
and promising performance was achieved by
`intermediate' regularization strategies between the sparse ($\ell_1$)
and dense ($\ell_2$) regularizers, e.g., 
a mixture of $\ell_1$-mixed-norm and $\ell_2$-mixed-norm
called the \emph{elastic-net regularization}
 \citep{NIPSWS:Taylor:2008,NIPSWS:ElastMKL:2009}
and $\ell_p$-mixed-norm regularization with $1<p<2$
\citep{JMLR:MicchelliPontil:2005,NIPS:Marius+etal:2009}.

Together with the active development of practical MKL optimization algorithms,
theoretical analysis of MKL has also been extensively conducted.
% While several methods related to MKL have been proposed, 
For $\ell_1$-mixed-norm MKL, 
\citet{COLT:Koltchinskii:2008} established the learning rate
$d^{\frac{1-s}{1+s}} n^{-\frac{1}{1+s}} + {d \log(M)}/{n}$
under rather restrictive conditions,
where $n$ is the number of samples, 
$d$ is the number of non-zero components of the ground truth,
$M$ is the number of kernels,
and $s$ ($0<s<1$) is a constant representing the complexity 
of the reproducing kernel Hilbert spaces (RKHSs).
Their conditions include a smoothness assumption of the ground truth ($q=1$ in our terminology (Assumption \ref{ass:convolution})).
For elastic-net regularization,
\cite{AS:Meier+Geer+Buhlmann:2009} gave a near optimal convergence rate
$d \left(n/\log(M)\right)^{-\frac{1}{1+s}}$.
Recently, \cite{AS:Koltchinskii+Yuan:2010}
showed that MKL with a variant of $\ell_1$-mixed-norm regularization 
achieves the minimax optimal convergence rate,
which successfully got a sharper dependency with respect to $\log(M)$ 
than the bound of \cite{AS:Meier+Geer+Buhlmann:2009} and
established the bound $d n^{-\frac{1}{1+s}} + {d \log(M)}/{n}$.
Another line of research
considers the cases where the ground truth is not sparse,
and bounds the Rademacher complexity of a candidate kernel class
by a pseudo-dimension of the kernel class
\citep{COLT:Srebro+BenDavid:2006,COLT:Ying+Campbell:2009,UAI:Cortes+etal:2009,ECML:Marius+etal:2010}.
% However, these works do not utilize the sparsity of the truth,
% and thus are not suited to sparse settings. 
%Moreover their order is $1/\sqrt{n}$ with respect to the number of samples
%because they do not utilize the {\it localization techniques}.

In this paper, we focus on the sparse setting
(i.e.,  the total number of kernels is large,
but the number of non-zero components of the ground truth is relatively small),
and derive a sharp learning rate for elastic-net MKL.
Our new learning rate,
$$
d^{\frac{1+q}{1+q+s}} n^{-\frac{1+q}{1+q+s}} R_{2,\gstar}^{\frac{2s}{1+q+s}} + \frac{d\log(M)}{n},
$$
is faster than all the existing bounds,
where $R_{2,\gstar}$ is a kind of the $\ell_2$-mixed-norm of the truth
and $q$ ($0 \leq q \leq 1$) is a constant depending on the smoothness of the ground truth.

% We suppose the ground truth is sparse,
% and the resultant bond is suited to sparse settings.

Our contributions are summarized as follows.
\begin{itemize}
\item 
% Our main notion is that,
The sharpest existing bound given by \citet{AS:Koltchinskii+Yuan:2010}
achieves the minimax rate on the $\ell_{\infty}$-mixed-norm ball \citep{NIPS:Raskutti+Martin:2009,arXiv:Raskutti+Martin:2010}. 
Our work follows this line and show that the learning rate for elastic-net MKL
further achieves the minimax rate on the \emph{$\ell_2$-mixed-norm ball},
which is faster than that on the $\ell_{\infty}$-mixed-norm ball.
This result implies that the bound by \citet{AS:Koltchinskii+Yuan:2010} is tight
only when the ground truth is evenly spread in the non-zero components.

\item 
We included the \emph{smoothness} $q$ of the ground truth into our learning rate,
where the ground truth is said to be smooth
if it is represented as a convolution of a certain function and an integral kernel (see Assumption \ref{ass:convolution}). 
Intuitively for larger $q$, the truth is smoother.
We show that, the smoother the truth is, the faster the convergence rate is.
That is, the resultant convergence rate becomes as if the complexity of RKHSs was $\frac{s}{1+q}$ 
instead of the true complexity $s$.
\citet{AS:Meier+Geer+Buhlmann:2009,AS:Koltchinskii+Yuan:2010} assumed $q=0$ and
\citet{COLT:Koltchinskii:2008} considered a situation of $q=1$. 
Our analysis covers those situations.
\end{itemize}

\section{Preliminaries}
In this section, we formulate elastic-net MKL,
and summarize mathematical tools that are needed for theoretical analysis.

\subsection{Formulation}
Suppose we are given $n$ samples $(x_i,y_i)_{i=1}^n$ where $x_i$ belongs to an input space $\calX$ and $y_i \in \Real$.
We denote the marginal distribution of $X$ by $\Pi$. 
We consider a MKL regression problem 
in which the unknown target function is represented as a form of 
$f(x)= \sum_{m=1}^M f_m(x)$ 
where each $f_m$ belongs to a different RKHS
$\calH_m~(m = 1,\dots,M)$ with kernel $k_m$ over $\calX\times\calX$.

The elastic-net MKL we consider in this paper is the version considered in \cite{AS:Meier+Geer+Buhlmann:2009}:
\begin{align}
\fhat = &\mathop{\arg \min}_{f_m \in \calH_m \atop (m=1,\dots,M)}
\frac{1}{n}\sum_{i=1}^N \left(y_i\!-\! \sum_{m=1}^M f_m(x_i) \right)^2\!\!\! +
\sum_{m=1}^M  \lambdaone \sqrt{ \|f_m\|_n^2 + \lambdatwo \|f_m\hnorm{m}^2 }  + \lambdathree \sum_{m=1}^M \|f_m\hnorm{m}^2,\;\;
\label{eq:primalElasticMKLnonpara}
\end{align}
where $\|f_m\|_n := \sqrt{ \frac{1}{n} f_m(x_i)^2}$ and $\|f_m\hnorm{m}$ is the RKHS norm of $f_m$ in $\calH_m$.
The regularizer is the mixture of $\ell_1$-term $\sum_m \sqrt{ \|f_m\|_n^2 + \lambdatwo \|f_m\hnorm{m}^2 }$ and $\ell_2$-term $\sum_m \|f_m\hnorm{m}^2$.
In that sense, we say that the regularizer is of the elastic-net type\footnote{
There is another version of MKL with elastic-net regularization considered in \cite{NIPSWS:Taylor:2008} and \cite{NIPSWS:ElastMKL:2009},
that is,
$ %\begin{align}
\lambdaone \sum_{m=1}^M \|f_m\hnorm{m} + \lambdatwo \sum_{m=1}^M \|f_m\hnorm{m}^2
$ %\end{align}
(i.e., there is no $\|f_m\|_n$ term in the regularizer). 
However, we focus on the former one because 
the later one is too loose to properly bound the irrelevant components of the estimated function. % and the truth on the irrelevant components. 
%that is more preferable in sparse settings from theoretical point of view. 
%In fact, our formulation achieves the minimax optimal rate for sparse ground truth.
} \citep{JRSS:Zou+Hastie:2005}.
Here the $\ell_1$ term is a mixture of the empirical $L_2$ norm $\|f_m\|_n$ and the RKHS norm $\|f_m\hnorm{m}$.
\cite{AS:Koltchinskii+Yuan:2010} also considered $\ell_1$ regularization that is a mixture of these quantities: $\sum_m \lambdaone \|f_m\|_n + \lambdatwo \|f_m\hnorm{m}$.

By the representer theorem \citep{JMAA:KimeldorfWahba:1971},
the solution $\fhat$ can be expressed as a linear combination of $nM$ kernels:
%\begin{equation}
$\exists \alpha_{m,i}\in \Real~,~ \fhat_m(x) = \sum_{i=1}^n \alpha_{m,i}k_m(x,x_i).$
%\end{equation}
Thus, using the Gram matrix $\boldK_m = (k_m(x_i,x_j))_{i,j}$,
the regularizer in \eqref{eq:primalElasticMKLnonpara} is expressed as 
\begin{align*}
\textstyle
\sum_{m=1}^M  \lambdaone \sqrt{ \boldalpha_m^\top\left(\frac{\boldK_m \boldK_m}{n} +\lambdatwo \boldK_m\right)\boldalpha_m} 
   + \lambdathree \sum_{m=1}^M \boldalpha_m^\top \boldK_m \boldalpha_m,
\end{align*}
where $\boldalpha_m = (\alpha_{m,i})_{i=1}^n \in \Real^n$.
Thus, we can solve the problem by a SOCP (second-order cone programming) solver as in \cite{ICML:Bach+etal:2004},
or the coordinate descent algorithms \citep{JRSS:Meier+etal:2008}.

\subsection{Notations and Assumptions}

Here, we present several assumptions used in our theoretical analysis and prepare notations.

Let $\totH = \calH_1 \oplus \dots \oplus \calH_M$. %= \{\sum_{m=1}^M f_m \mid f_m \in \calH_m \}$.  
We denote by $\fstar \in \totH$ the ground truth satisfying the following assumption.
%Throughout the paper, we assume the following technical conditions 
%(see also \cite{JMLR:BachConsistency:2008}). 

\begin{Assumption}{\bf(Basic Assumptions)}\ 
\label{ass:basic}
\begin{enumerate}
\item[{\rm (A\ref{ass:basic}-1)}] %$\mathrm{(1)}$]
There exists $\fstar = (\fstar_1,\dots,\fstar_M) \in \totH$
such that $\EE[Y|X] = \sum_{m=1}^M \fstar_m(X)$,
and the noise $\epsilon := Y - \fstar(X)$ 
%has a strictly positive variance;
%there exists $\sigma>0$ such that $\EE[\epsilon^2 | X] > \sigma^2 $ for all $X \in \calX$.
%We also assume that $\epsilon$ 
is bounded as $|\epsilon| \leq L$.
\item[{\rm (A\ref{ass:basic}-2)}]
For each $m=1,\dots,M$, $\calH_m$ is separable and $\sup_{X\in \calX} |k_m(X,X)| \leq 1$.
\end{enumerate}
\end{Assumption}
%\textbf{SUGIYAMA======Can you briefly explain the meaning of assumption (A1)?============}
The first assumption in (A\ref{ass:basic}-1) ensures the model $\calH$ is correctly specified, 
and the technical assumption $|\epsilon| < L$ allows $\epsilon f$ to be Lipschitz continuous with respect to $f$.
These assumptions are not essential and can be relaxed to misspecified models and unbounded noise such as Gaussian noise \citep{arXiv:Raskutti+Martin:2010}.
 %~\cite{JMLR:BachConsistency:2008,FCM:Caponetto+Vito:2007}, but
However, for the sake of simplicity, 
we assume these conditions.

It is known that the assumption (A\ref{ass:basic}-2) gives the following relation: % between the $\ell_\infty$-norm and the RKHS norm:
\begin{align*}
\|f_m\|_{\infty} \!\! \leq \! \sup_{x}\langle k_m(x,\cdot), f_m \rangle_{\calH_m}
%&
\!\! \leq \! \sup_{x} \| k_m(x,\cdot)\hnorm{m}\! \|f_m\hnorm{m} %\\
%& 
\!\! \leq \!  
\sup_{x}  \sqrt{k_m(x,x)} \|f_m\hnorm{m} 
\! \leq \! \|f_m\hnorm{m}.
\end{align*}
% The assumption (A3) ensures that $\fstar$ is unique.
Later, we will also assume a stronger (but practical) condition on the sup-norm in Assumption \ref{ass:supnorm}.

%\begin{table}[t]
%\centering
%\caption{Summary of the constants we use in this article.}
%\label{tab:constants}
%\begin{tabular}{|c|l|}
%\hline
%$M$ & The number of candidate kernels.  \\ \hline
%$d$ & The number of active kernels of the truth; i.e., $d=|I_0|$. \\ \hline
%$R$ & The upper bound of $\sum_{m=1}^M (\|\fstar_m \hnorm{m} +
%     \|\fstar_m \hnorm{m}^2)$; see (A4). \\ \hline
%$s$ & The spectral decay coefficient; see (A5). \\ \hline 
%\end{tabular}
%\end{table}

%We define an operator $\Sigma_{m,m'}:\calH_{m'} \to \calH_{m}$ as 
%$$
%\langle f_m,\Sigma_{m,m'} g_m' \rangle := \EE[f_m(X) g_{m'}(X)].
%$$
%In particular, we denote $\Sigma_{m,m}$ by $T_m$:
We define an operator $T_{m}:\calH_{m} \to \calH_{m}$ as 
$$
\langle f_m,T_{m} g_m \rangle_{\calH_m} := \EE[f_m(X) g_{m}(X)],
$$
where $f_m,g_m \in \calH_m$.
Due to Mercer's theorem, 
there are an orthonormal system $\{\phi_{k,m}\}_{k,m}$ in $L_2(\Pi)$
and the spectrum $\{\mu_{k,m}\}_{k,m}$
such that $k_m$ has the following spectral representation: %is represented as 
%the spectral representation: % the kernel function $k_m$ 
\begin{equation}
k_m(x,x') = \sum_{k=1}^{\infty} \mu_{k,m} \phi_{k,m}(x) \phi_{k,m}(x'). 
\label{eq:spectralRepre}
\end{equation}
By this spectral representation, the inner-product of RKHS can be expressed as 
$
\langle f_m ,g_m \rangle_{\calH_m} = \sum_{k=1}^{\infty} \mu_{k,m}^{-1} \langle f_m, \phi_{k,m} \rangle_{\LPi} \langle \phi_{k,m}, g_m \rangle_{\LPi}.
$

\begin{Assumption}{\bf (Convolution Assumption)}
\label{ass:convolution}
There exist a real number $0 \leq q \leq 1$ and $\gstar_m \in \calH_m$ such that 
\begin{flalign*}
\text{\rm(A\ref{ass:convolution})} &&
\fstar_m(x) = \int_{\calX}k_m^{(q/2)}(x,x')\gstar_m(x')\dd \Pi(x')
 \qquad(\forall m = 1,\dots,M), && %\label{eq:fstarSigmacond}
\end{flalign*}
where $k_m^{(q/2)}(x,x')= \sum_{k=1}^{\infty} \mu_{k,m}^{q/2}
  \phi_{k,m}(x) \phi_{k,m}(x')$. 
This is equivalent to the following operator representation:
$$
\fstar_m = T_m^{\frac{q}{2}} \gstar_m.
$$
\end{Assumption}
The constant $q$ controls the smoothness of the truth $\fstar_m$
because $\fstar_m$ is a convolution of the integral kernel $k_m^{(q/2)}$ and $\gstar_m$,
and high frequency components are depressed as $q$ becomes large. 
Therefore, as $q$ becomes large, $\fstar$ becomes ``smooth''. 
The assumption (A\ref{ass:convolution}) was considered in \cite{FCM:Caponetto+Vito:2007} to analyze 
the convergence rate of least-squares estimators in a single kernel setting. 
In MKL settings,  
\cite{COLT:Koltchinskii:2008} showed a fast learning rate of MKL,
and \cite{JMLR:BachConsistency:2008} employed the assumption for $q=1$
to show the consistency of MKL.
%It ensures the consistency of the least-squares estimates in terms of the RKHS norm.  
%This condition (for $q=1$) was also assumed in \cite{COLT:Koltchinskii:2008}. 
Proposition 9 of \cite{JMLR:BachConsistency:2008} gave
a sufficient condition to fulfill (A\ref{ass:convolution}) with $q=1$ for translation invariant kernels $k_m(x,x') = h_m(x-x')$.
\cite{AS:Meier+Geer+Buhlmann:2009} considered a situation with $q=0$ on Sobolev space;
the analysis of \cite{AS:Koltchinskii+Yuan:2010} also corresponds to $q=0$.
Note that (A\ref{ass:convolution}) with $q=0$
imposes nothing on the smoothness about the truth,
and our analysis also covers this case.

We will show in Appendix~\ref{appendix:CoveringNumber} that 
as $q$ increases, the space of the functions that satisfy (A\ref{ass:convolution}) becomes ``simple''.
Thus, it might be natural to consider that, under the Convolution Assumption (A\ref{ass:convolution}),
the learning rate becomes faster as $q$ increases. 
Although this conjecture is actually true,
it is not obvious because the Convolution Assumption only restricts the ground truth,
but not the search space.

%(Technically, $\|\fstar_m \hnorm{m} - \|f_m\hnorm{m} \leq \| \fstar_m - f_m \|_{\LPi} \frac{\|\gstar_m\hnorm{m}}{\|\fstar_m\hnorm{m}}$
%(see \eqref{eq:secondboundforbasic}) gives a sharp convergence bound)
%Using the spectral representation \eqref{eq:spectralRepre},
%the condition $\gstar_m \in \calH_m$ is expressed as    
%\begin{equation}
%\|\gstar_m\hnorm{m}^2 =\sum_{k=1}^{\infty} \mu_{k,m}^{-(1+q)} \langle \fstar_m,\phi_{k,m} \rangle_{\LPi}^2 < \infty.
%\end{equation}

%\begin{Assumption}{\bf (Boundedness Assumption)}
%\label{ass:bounded}
%There exists a constant $R$ such that 
%\begin{flalign*}
%\text{\rm(A\ref{ass:bounded})} & & \|\gstar_m\hnorm{m}\leq R, &&
%\end{flalign*}
%for all $m$.
%\end{Assumption}

Next we introduce a parameter representing the complexity of RKHSs.
\begin{Assumption}{\bf (Spectral Assumption)}
\label{ass:specass}
There exist $0 < s < 1$ and $c$ such that %the spectrum $\mu_{k,m}$ has the following decreasing exponent
\begin{flalign*}
%\label{eq:spetrumassump}
\text{\rm(A\ref{ass:specass})} &&
\mu_{k,m} \leq c k^{-\frac{1}{s}},~~~(1\leq \forall k, 1\leq \forall m \leq M),&&
\end{flalign*}
where $\{\mu_{k,m}\}_{k}$ is the spectrum of the kernel $k_m$ (see Eq.\eqref{eq:spectralRepre}).
\end{Assumption}
It was shown that the spectral assumption (A\ref{ass:specass}) is equivalent to 
the classical covering number assumption\footnote{
The $\epsilon$-covering number $\calN(\epsilon,\mathcal{B}_{\calH_m},\LPi)$ with respect to $\LPi$
is the minimal number of balls with radius $\epsilon$ needed to cover the unit ball $\mathcal{B}_{\calH_m}$ in $\calH_m$ \citep{Book:VanDerVaart:WeakConvergence}.
} \citep{COLT:Steinwart+etal:2009}.
If the spectral assumption (A\ref{ass:specass}) holds, there exists a constant $C$ that
depends only on $s$ and $c$ such that 
\begin{align}
\label{eq:coveringcondition}
\calN(\varepsilon,\mathcal{B}_{\calH_m},\LPi) \leq C \varepsilon^{-2 s},
\end{align}
and the converse is also true (see Theorem 15 of \cite{COLT:Steinwart+etal:2009} and \cite{Book:Steinwart:2008} for details).
Therefore, 
if $s$ is large, 
at least one of the RKHSs is ``complex'',
and if $s$ is small, all the RKHSs are ``simple''.
A more detailed characterization of the covering number
in terms of the spectrum is provided in Appendix~\ref{appendix:CoveringNumber}.
The covering number of the space of functions that satisfy 
the Convolution Assumption (A\ref{ass:convolution}) is also provided there.

%Similarly, we define $\Sigma_{I,J}$ and $\VCor_{I,J}$ as the restriction to an index set $I \times J$ for $I,J \subseteq \{1,\dots,M\}$. 
We denote by $\Istar$ the indices of truly active kernels, i.e.,
$$
\Istar :=\{m \mid \|\fstar_m\hnorm{m}>0\}.
$$
%and define the complement of $\Istar$ as $\Jstar = {\Istar}^c$.
For $f =\sum_{m=1}^M f_m \in \totH$ 
and a subset of indices $I \subseteq \{1,\dots,M\}$, we define $\calH_I = \oplus_{m\in I} \calH_m$ and 
denote by $f_I \in \calH_I$ the restriction of $f$ to an index set $I$, i.e., $f_I = \sum_{m \in I} f_m$. 
For a given set of indices $I \subseteq \{1,\dots, M \}$, % and $J = I^c$,
let $\kmin(I)$ be defined as follows:
\begin{align*}
\kmin(I) &:= \sup\left\{\kappa \geq 0 ~\Big|~ \kappa \leq 
\frac{\|\sum_{m\in I}f_m\|_{\LPi}^2}{\sum_{m\in I}\|f_m\|_{\LPi}^2} ,~\forall f_m \in \calH_m~(m\in I)\right\}. %,
\end{align*}
$\kappa(I)$ represents the correlation of RKHSs inside the
 indices $I$.  Similarly, we define the {\it canonical correlations} of RKHSs between $I$ and $I^c$ as follows:
\begin{align*}
\rho(I) &:= \sup \left\{\frac{\langle f_I, g_{I^c} \rangle_{\LPi} }{\|f_I\|_{\LPi}\|g_{I^c} \|_{\LPi}} 
~\Big|~ f_I \in \calH_I, g_{I^c} \in \calH_{I^c}, f_I \neq 0, g_{I^c} \neq 0 \right\}. 
\end{align*}
These quantities give a connection between the $\LPi$-norm of $f \in \totH$ and the $\LPi$-norm of $\{f_m\}_{m\in I}$
as shown in the following lemma. The proof is given in Appendix \ref{sec:appendixLemm}.
\begin{Lemma}
\label{lem:incoherenceIneq}
For all $I \subseteq \{1,\dots,M\}$, we have
\begin{align*}
\| f \|_{\LPi}^2 \geq  (1- \rho(I)^2) \kmin(I) \left(\sum_{m \in I}\| f_m  \|_{\LPi}^2\right).
\end{align*}
\end{Lemma}
We impose the following assumption for $\kappa(I_0)$ and $\rho(I_0)$.
\begin{Assumption}{\bf (Incoherence Assumption)}
\label{ass:incoherence}
For the truly active components $I_0$, $\kmin(\Istar)$ is strictly positive and $\rho(I_0)$ is strictly less than 1:
\begin{flalign}
%\rho(I) < 1.
%\frac{\kmax(\Istar)}{\kmin(\Istar)(1-\rho^2(\Istar))} < C_3.
\text{\rm(A\ref{ass:incoherence})} && 0 < \kmin(\Istar)(1-\rho^2(\Istar)). &&
%\label{eq:NoPerfectCorrelationOne}
\end{flalign}
\end{Assumption}

%For the truly active components $I_0$, 
%\begin{flalign}
%&& 0 < \kmin(\Istar)(1-\rho^2(\Istar)). &&
%\end{flalign}
 
This condition is known as the {\it incoherence condition} \citep{COLT:Koltchinskii:2008,AS:Meier+Geer+Buhlmann:2009},
i.e., RKHSs are not too dependent on each other. % and the problem is well conditioned.
In the theoretical analysis, we also obtain an upper bound of the $\LPi$-norm of $\fhat - \fstar$ in terms of the $\LPi$-norm of $\{\fhat_m - \fstar_m\}_{m\in I_0}$.
Thus, by the incoherence condition and Lemma \ref{lem:incoherenceIneq}, 
we may focus on bounding the $\LPi$-norm of the ``low-dimensional'' components
$\{\fhat_m - \fstar_m\}_{m\in I_0}$, instead of all the components.
\cite{AS:Koltchinskii+Yuan:2010} considered a weaker condition
including the \emph{restricted isometry} \citep{AS:Candes+Tao:2007} instead of (A\ref{ass:incoherence}).
Such a weaker condition is also applicable to our analysis, 
but we employ (A\ref{ass:incoherence}) for simplicity.

Finally we impose the following technical assumption related to the sup-norm of the members in the RKHSs.
\begin{Assumption}{\bf (Sup-norm Assumption)}
\label{ass:supnorm}
Along with the Spectral Assumption (A\ref{ass:specass}),  
there exists a constant $C_1$ such that 
\begin{flalign*}
\text{\rm(A\ref{ass:supnorm})} &&
\|f_m\|_{\infty} \leq C_1 \|f_m\|_{\LPi}^{1-s}\|f_m\hnorm{m}^s~~~(\forall f_m \in \calH_m,m=1,\dots,M),&&
\end{flalign*}
where $s$ is the exponent defined in the Spectral Assumption (A\ref{ass:specass}).
\end{Assumption}
This assumption is satisfied if the RKHS is a Sobolev space or 
is continuously embeddable in a Sobolev space. 
For example, the RKHSs of Gaussian kernels are continuously embedded in all Sobolev spaces,
and thus satisfy the Sup-norm Assumption (A\ref{ass:supnorm}).
More generally, RKHSs with $m$-times continuously differentiable kernels on a closed Euclidean ball in $\Real^d$ are also 
continuously embedded in a Sobolev space, and satisfy the Sup-norm Assumption (A\ref{ass:supnorm}) with $s=\frac{d}{2m}$ (see Corollary 4.36 of \cite{Book:Steinwart:2008}).
Therefore this assumption is somewhat common for practically used kernels.
A more general necessary and sufficient condition in terms of {\it real interpolation} is shown in \cite{Book:Bennett+Sharpley:88}.
\cite{COLT:Steinwart+etal:2009} used this assumption to show the optimal rates for regularized regression using a single kernel function, 
and 
one can find detailed discussions about the assumption there.

%Constants we use later are summarized in Table~\ref{tab:constants}.

\section{Convergence rate analysis}
In this section, we present our main result.

\subsection{The convergence rate of elastic-net MKL}
Here we derive the learning rate of the estimator $\fhat$
defined by \Eqref{eq:primalElasticMKLnonpara}. 
We denote the number of truly active components by $d := |\Istar |$.
We may suppose that the number of kernels $M$ and
the number of active kernels $d$ are increasing
with respect to the number of samples $n$.
Our main purpose of this section is to show that the learning rate can be 
faster than the existing bounds.
The existing bound has already been shown to be optimal on the $\ell_{\infty}$-mixed-norm ball \cite{AS:Koltchinskii+Yuan:2010,arXiv:Raskutti+Martin:2010}. 
Our claim is that the convergence rate can further achieve 
the minimax optimal rate on the \emph{$\ell_2$-mixed-norm ball},
which is faster than that on the $\ell_{\infty}$-mixed-norm ball. 

%First we derive the learning rate.

Define $\eta(t)$ for $t>0$ as 
\[
\eta(t) := \max(1,\sqrt{t},t/\sqrt{n}).
\]
For given $\lambdatmp > 0$, we define $\xi_n$ as 
\begin{equation}
\xi_n := \xi_n(\lambdatmp) = \left( \frac{\lambdatmp^{-\frac{s}{2}}}{\sqrt{n}} \vee \frac{\lambdatmp^{-\frac{1}{2}}}{n^{\frac{1}{1+s}}} \vee \sqrt{\frac{\log(M)}{n}}\right).
\label{eq:definitionxin}
\end{equation}

\begin{Theorem}
\label{eq:TheConvergenceRateMain}
Suppose Assumptions \ref{ass:basic}--\ref{ass:supnorm} are satisfied, and let $\lambdatmp >0$ be an arbitrary positive number.
Then there exist universal constants $\tilde{C}_1,\tilde{C}_2$ and a constant $\psi_s$ depending on $s,c,L,C_1$ 
such that 
if $\lambdaone$, $\lambdatwo$ and $\lambdathree$ are set as 
$\lambdaone = \psi_s \eta(t) \xi_n(\lambdatmp)$, $\lambdatwo = \lambdatmp$, $\lambdathree = \lambdatmp$, 
then for all $n$ and $r(>0)$ satisfying $\frac{\log(M)}{\sqrt{n}} \leq 1$ and the inequality %that satisfies  
\begin{align}
%\frac{\phi_s C_4^2}{(1-\rho(I_0))^2 \kappa(I_0)}d \sqrt{n} \xi_n^2 < \frac{1}{8},
\frac{\tilde{C}_1 \max(\psi_s \sqrt{n} \xi_n^2,r) \left( d + \frac{\lambdathree^{1+q}}{\lambdaone^2}\sum_{m=1}^M\|\gstar_m\hnorm{m}^2  \right)}{(1- \rho(I_0)^2) \kmin(I_0)} \leq 1,
\label{eq:ForLargenMain}
\end{align}
we have 
\begin{align}
\|\fhat - \fstar \|_{\LPi}^2 \leq \frac{\tilde{C}_2}{(1-\rho(I_0))^2 \kappa(I_0)} \left(d  \lambdaone^2 +  \lambdathree^{1+q}  \sum_{m=1}^M \|\gstar_m \hnorm{m}^2 \right), 
\label{eq:TheBoundMain}
\end{align}
with probability $1- \exp(- t) - \exp\left(-\min\left\{ \frac{r^2 \log(M)}{n\xi_n(\lambdatmp)^4\psi_s^2}, \frac{r}{\xi_n(\lambdatmp)^2\psi_s}\right\}\right)$ for all $t \geq 1$.
\end{Theorem}

A proof of Theorem~\ref{eq:TheConvergenceRateMain} is provided in
Appendix~\ref{proof:TheConvergenceRateMain}.
The convergence rate \eqref{eq:TheBoundMain} contains a tuning parameter $\lambdatmp$. 
Here we optimize this parameter.
Let 
\begin{equation*}
R_{p,\gstar} := \left(\sum_{m=1}^M \|\gstar_m \hnorm{m}^p \right)^{\frac{1}{p}},
\end{equation*}
and we assume that $R_{p,\gstar}$ is strictly positive for all $p\geq 1$ ($R_{p,\gstar} > 0$).
If $n$ is sufficiently large compared with $R_{2,\gstar}$, the RHS of \Eqref{eq:TheBoundMain} is minimized by 
\begin{equation*}
\lambdatmp = d^{\frac{1}{1+q+s}} n^{-\frac{1}{1+q+s}}R_{2,\gstar}^{-\frac{2}{1+q+s}}, 
\end{equation*}
up to constants. 
Then the convergence rate  \eqref{eq:TheBoundMain} is reduced to 
\begin{align}
\|\fhat - \fstar \|_{\LPi}^2 \leq \widehat{C}_1 \! \left( d^{\frac{1+q}{1+q+s}}n^{-\frac{1+q}{1+q+s}}R_{2,\gstar}^{\frac{2s}{1+q+s}} 
+ \frac{d \log(M)}{n} + d^{\frac{q+s}{1+q+s}} n^{-\frac{1+q}{1+q+s} - \frac{q(1-s)}{(1+s)(1+q+s)}}R_{2,\gstar}^{\frac{2}{1+q+s}} \right),~~
\label{eq:roughbound1}
\end{align}
where $\widehat{C}_1$ is a constant.
If $n^{\frac{q}{1+s}} \frac{d}{R^2_{2,\gstar}} \geq C$ with a constant $C$ (this holds if $\|\gstar_m\hnorm{m} \leq \sqrt{C}$ for all $m$),
then \Eqref{eq:roughbound1} becomes 
\begin{align}
\|\fhat - \fstar \|_{\LPi}^2 \leq \widehat{C}_2  \left( d^{\frac{1+q}{1+q+s}}n^{-\frac{1+q}{1+q+s}}R_{2,\gstar}^{\frac{2s}{1+q+s}} + \frac{d \log(M)}{n} \right),
\label{eq:roughbound2}
\end{align}
where $\widehat{C}_2$ is a constant.
We see that, as $q$ becomes large (the truth becomes smooth) or $s$ becomes small (the RKHSs become simple), 
the convergence rate becomes faster when $R_{2,\gstar}\geq 1$. 
In the next subsection, we show that this bound \eqref{eq:roughbound2} achieves the minimax optimal rate on the $\ell_2$-mixed-norm ball.

\subsection{Minimax learning rate of $\ell_2$-mixed-norm ball}
To derive the minimax rate, we slightly simplify the setup.
First, we assume that the input $\calX$ is
expressed as $\calX = \tilde{\calX}^M$ for some space $\tilde{\calX}$.
%  an $M$ product of a space $\tilde{\calX}$, i.e., 
Second, all the RKHSs $\{\calH_m\}_{m=1}^M$ are the same as
an RKHS $\tilde{\calH}$ defined on $\tilde{\calX}$.
Finally, we assume that the marginal distribution $\Pi$ of input is a product of a probability distribution $Q$, i.e., $\Pi = Q^M$.
Thus, an input $x=(\tilde{x}^{(1)},\dots,\tilde{x}^{(M)}) \in \calX = \tilde{\calX}^M$ is a concatenation of $M$ random variables $\{\tilde{x}^{(m)}\}_{m=1}^M$
independently and identically distributed from the distribution $Q$.
Moreover, the function class $\totH$ is a class of functions $f$ such that  
$$
f(x) = f(\tilde{x}^{(1)},\dots,\tilde{x}^{(M)}) =  \sum_{m=1}^M f_m(\tilde{x}^{(m)}),
$$ 
where $f_m \in \tilde{\calH}$ for all $m$.
Without loss of generality, we may assume that all functions in $\tilde{\calH}$ are centered:
$$
\EE_{\tilde{X} \sim Q}[f(\tilde{X})] = 0~~~(\forall f \in \tilde{\calH}).
$$
We assume that the spectrum of the kernel $\tilde{k}$ corresponding to the RKHS $\tilde{\calH}$ decays at the rate of $-\frac{1}{s}$.
That is,
in addition to Assumption \ref{ass:specass}, we impose the following lower bound to the spectrum: % there exists $0 < s < 1$ such that %and $c_1,c_2$ such that 
there exist $c',c$ $(>0)$ such that 
\begin{flalign}
%\text{\rm(A\ref{ass:specass})} 
&&
c'k^{-\frac{1}{s}} \leq \mu_{k} \leq c k^{-\frac{1}{s}},&&
\label{eq:strongSpecAss}
\end{flalign}
where $\{\mu_{k}\}_{k}$ is the spectrum of the kernel $\tilde{k}$ (see Eq.\eqref{eq:spectralRepre}).
We also assume that the noise $\{\epsilon_i\}_{i=1}^n$ is generated by a Gaussian distribution with mean 0 and standard deviation $\sigma$.

Let $\calH_{\ell_0}(d)$ be the set of functions with $d$ non-zero components in $\totH$ 
defined by
\begin{align*}
\calH_{\ell_0}(d) := \{ (f_1,\dots,f_M) \in \totH \mid |\{m \mid \|f_m\hnorm{m} \neq 0\}| \leq d  \}.
\end{align*}
We define $\ell_p$-mixed-norm ball ($p\geq 1$) with radius $R$ in $\calH_0(d)$ as
\begin{align*}
\calH_{\ell_p}^{d,q}(R) := \left\{ f = \sum_{m=1}^M f_m ~\Big|~ \exists (g_1,\dots,g_M) \in \calH_0(d),~f_m = T_m^{\frac{q}{2}} g_m,~ \left(\textstyle \sum_{m=1}^M \|g_m\hnorm{m}^p\right)^{\frac{1}{p}} \leq R
  \right\}.
\end{align*}
In \cite{arXiv:Raskutti+Martin:2010}, the minimax learning rate on $\calH_{\ell_{\infty}}^{d,0}(R)$ (i.e., $p=\infty$ and $q=0$) was derived\footnote{
The set $\calF_{M,d,\calH}(R)$ in \cite{arXiv:Raskutti+Martin:2010}
corresponds to $\calH_{\ell_{\infty}}^{d,0}(R)$ in the current paper.
}.
We show (a lower bound of) the minimax learning rate for more general settings
($p=2,\infty$ and $0\leq q \leq 1$)
in the following theorem.

\begin{Theorem}
\label{eq:LowerboundOfElastMKL}
Let $\sprime = \frac{s}{1+q}$. Assume $d \leq M/4$.
Then the minimax learning rates are lower bounded as follows.
There exists a constant $\tilde{C}_1$ such that for $R_2\geq\sqrt{\frac{d\log(M/d)}{n}}$, the radius of the $\ell_2$-mixed-norm ball, we have 
\begin{align}
\inf_{\fhat} \sup_{\fstar \in \calH_{\ell_2}^{d,q}(R_2)}\EE[\|\fhat - \fstar \|_{\LPi}^2] 
\geq \tilde{C}_1 \left( d^{\frac{1}{1+\sprime}} n^{-\frac{1}{1+\sprime}}R_2^{\frac{2 \sprime}{1+\sprime}} + \frac{d \log(M/d)}{n} \right),
\label{eq:minimaxL2}
\end{align}
where `inf' is taken over all measurable functions of the samples $(x_i,y_i)_{i=1}^n$ and the expectation is taken for the sample distribution. 
Similarly, we have the following minimax-rate for $p=\infty$:
\begin{align}
\inf_{\fhat} \sup_{\fstar \in \calH_{\ell_{\infty}}^{d,q}(R_{\infty})}\EE[\|\fhat - \fstar \|_{\LPi}^2] 
\geq \tilde{C}_1 \left( d n^{-\frac{1}{1+\sprime}}R_{\infty}^{\frac{2 \sprime}{1+\sprime}} + \frac{d \log(M/d)}{n} \right),
\label{eq:minimaxLinf}
\end{align}
for $R_{\infty}\geq\sqrt{\frac{\log(M/d)}{n}}$.
\end{Theorem}

A proof of Theorem~\ref{eq:LowerboundOfElastMKL} is provided in
Appendix~\ref{proof:LowerboundOfElastMKL}.

Obviously, our learning rate \eqref{eq:roughbound2} of elastic-net MKL achieves the minimax optimal rate \eqref{eq:minimaxL2} on the $\ell_2$-mixed-norm ball if $M \gg d$.
Moreover, the optimal rate (9) on the $\ell_2$-mixed-norm ball is always
faster than that of $\ell_\infty$-mixed-norm (10).
%The optimal rate \eqref{eq:minimaxLinf} on the $\ell_\infty$-mixed-norm ball is always larger than that of the $\ell_2$-mixed-norm ball \eqref{eq:minimaxL2}.
To see this, let $R_{\infty,\gstar} := \max_{m} \|\gstar_m \hnorm{m}$;
then we always have $R_{2,\gstar} \leq \sqrt{d} R_{\infty,\gstar}$
and consequently we have 
$$d^{\frac{1}{1+\sprime}} n^{-\frac{1}{1+\sprime}}R_{2,\gstar}^{\frac{2 \sprime}{1+\sprime}} \leq d n^{-\frac{1}{1+\sprime}}R_{\infty,\gstar}^{\frac{2 \sprime}{1+\sprime}}.$$
Now we consider two examples, ``inhomogeneous setting'' and ``homogeneous setting'', to compare these two bounds:
\begin{enumerate}
\item $\|\gstar_m\hnorm{m} = m^{-1}$~($\forall m \in I_0=\{1,\dots,d\}$) (inhomogeneous setting): In this situation, $R_{\infty,\gstar} = 1$ and $R_{2,\gstar} \leq 1$. Thus,
the learning rate \eqref{eq:roughbound2} of elastic-net MKL and the minimax rate on the $\ell_2$-mixed-norm ball are $d^{\frac{1}{1+\sprime}} n^{-\frac{1}{1+\sprime}} + \frac{d\log(M)}{n}$
and that on the $\ell_\infty$-mixed-norm ball is $d n^{-\frac{1}{1+\sprime}} + \frac{d\log(M)}{n}$. 
Therefore, in the first term (the leading term with respect to $n$),
there is a difference in the $d^{\frac{\tilde{s}}{1+\tilde{s}}}$ factor.
This difference could be $\sqrt{d}$ in the worst case. 
Thus, there appears large discrepancy between the two rates in high-dimensional settings. %in the near sparse setting. 
\item $\|\gstar_m\hnorm{m} = 1$ ($\forall m \in I_0$) (homogeneous setting): 
In this situation, $R_{\infty,\gstar} = 1$ and $R_{2,\gstar} = \sqrt{d}$. Thus, all the bounds are 
$d n^{-\frac{1}{1+\sprime}} + \frac{d\log(M)}{n}$. Here we observe that the learning rate \eqref{eq:roughbound2} of elastic-net MKL coincides 
with the minimax rate on the $\ell_\infty$-mixed-norm ball.
We also notice that 
the homogeneous setting is the only situation where those two rates coincide with each other.
As seen later, the existing bounds by previous works are the minimax rate on the $\ell_\infty$-mixed-norm ball, thus are tight only in the homogeneous setting.
\end{enumerate}

\subsection{Comparison with existing bounds}
Here we compare the existing bounds and the bound we derived.
Roughly speaking, the difference from the existing bounds is summarized in the following two points:
\begin{itemize}
\item[(a)] Our learning rate achieves the minimax-rate of $\ell_2$-mixed-norm ball,
  instead of the $\ell_\infty$-mixed-norm ball. 
\item[(b)] Our bound includes the smoothing parameter $q$ (Assumption \ref{ass:convolution}),
  and thus is more general and faster than existing bounds.
\end{itemize}

The first bound on the convergence rate of MKL was derived by \cite{COLT:Koltchinskii:2008},
which assumed $q=1$ and $\frac{1}{d} \sum_{m \in I_0}\frac{\|\gstar_m\hnorm{m}^2}{\|\fstar_m\hnorm{m}^2} \leq C$. 
Under these rather strong conditions, they showed the bound $d^{\frac{1-s}{1+s}}n^{-\frac{1}{1+s}} + \frac{d\log(M)}{n}$.
For the smooth case $q=1$, we obtained a faster rate $n^{-\frac{2}{2+s}}$ instead of
$n^{-\frac{1}{1+s}}$ in their bound with respect to n.
%Unfortunately, the rate with respect to $n$ in the first term is $n^{-\frac{1}{1+s}}$,
%which is slower than $n^{-\frac{2}{2+s}}$ in our bound.

The second bound was given by \cite{AS:Meier+Geer+Buhlmann:2009}, which showed 
$d \left(\frac{\log(M)}{n}\right)^{\frac{1}{1+s}}$
for elastic-net regularization \eqref{eq:primalElasticMKLnonpara} under $q=0$.
Their bound almost achieves the minimax rate on the $\ell_\infty$-mixed-norm ball
except the additional $\log(M)$ term.
Compared with our bound, their bound has the $\log(M)$ term and the rate with respect to $d$
is larger than $d^{\frac{1}{1+s}}$ in our bound.

Most recently, \cite{AS:Koltchinskii+Yuan:2010} presented the bound $n^{-\frac{1}{1+s}}(d + \sum_{m\in I_0} \|\fstar_m\hnorm{m}) + \frac{d\log(M)}{n}$ for $q=0$.
Their bound is exactly the minimax rate on the $\ell_\infty$-mixed-norm ball.
However, their bound is $d^{\frac{s}{1+s}}$ times slower than ours if the ground truth is inhomogeneous.
For example, when $\|\fstar_m\hnorm{m} = m^{-1}$ ($m\in I_0 =\{1,\dots,d\}$) and $\fstar_m=0$ (otherwise), their bound is $n^{-\frac{1}{1+s}}d + \frac{d\log(M)}{n}$, 
while our bound is $n^{-\frac{1}{1+s}}d^{\frac{1}{1+s}} + \frac{d\log(M)}{n}$.

All the bounds explained above focused on either $q=0$ or $1$.
On the other hand, our analysis is more general in that
the whole range of $0\leq q \leq 1$ can be accommodated.

The relation between our analysis and existing analyses are summarized in Table \ref{tab:RelationOfBounds}.

\begin{table}[t]
\label{tab:RelationOfBounds}
\begin{center}
\caption{Relation between our analysis and existing analyses.}
\begin{tabular}{l|cccc}
        & regularizer &  smoothness ($q$)  & minimaxity   & convergence rate \\ \hline
%\cite{COLT:Koltchinskii:2008}    
K\&Y (2008) 
&   $\ell_1$          &    $q=1$            &     ?     &  $d^{\frac{1-s}{1+s}}n^{-\frac{1}{1+s}} + \frac{d\log(M)}{n}$ \\
\cite{AS:Meier+Geer+Buhlmann:2009}   &  Elastic-net     &    $q=0$           &    not optimal &  $d\left(\frac{\log(M)}{n}\right)^{\frac{1}{1+s}}$  \\
%\cite{AS:Koltchinskii+Yuan:2010}      
K\&Y (2010)&   variant of $\ell_1$          &    $q=0$        &    $\ell_\infty$-ball     &  $dn^{-\frac{1}{1+s}} + \frac{d\log(M)}{n}$ \\
This paper    & Elastic-net             &   $0\leq q\leq 1$         &    $\ell_2$-ball          & $\left(\frac{d}{n}\right)^{\frac{1+q}{1+q+s}} R_{2,\gstar}^{\frac{2s}{1+q+s}} + \frac{d\log(M)}{n}$ \\ \hline
\end{tabular}
\end{center}
\end{table}

\section{Conclusion and Discussion}
We presented a new learning rate of elastic-net MKL,
which is faster than the existing bounds of several MKL formulations.
According to our bound, 
the learning rate of elastic-net MKL achieves the minimax rate on the $\ell_2$-mixed-norm ball,
instead of the $\ell_\infty$-mixed-norm ball.
Our bound includes a parameter $s$ representing the complexity of the RKHSs and another parameter $q$ controlling the smoothness of the truth.
Under a natural condition, the learning rate becomes faster as $s$ becomes small or $q$ becomes large.
Although the existing works concluded that MKL is optimal in a sense that it achieves the minimax rate of the $\ell_\infty$-mixed-norm ball,
we presented that elastic-net MKL further achieves the minimax rate of the $\ell_2$-mixed-norm ball which is faster than that of the $\ell_\infty$-mixed-norm ball.

\cite{AS:Koltchinskii+Yuan:2010} considered a variant of $\ell_1$ regularization:
$\sum_{m=1}^M \lambdaone \|f_m\|_n + \lambdatwo \|f_m\hnorm{m}$.
They showed that MKL with that regularization achieves the minimax rate of the $\ell_\infty$-mixed-norm ball.
It might be interesting to investigate whether that regularization also achieves the minimax rate of the $\ell_2$-mixed-norm ball or another faster rate.
In particular, it is interesting to study whether the smoothness parameterization ($q$) gives a faster rate also for that $\ell_1$ regularization.
If not, that might explain the effectiveness of the elastic-net regularization in real data experiments. 

\appendix 
\section{Covering Number}
\label{appendix:CoveringNumber}
Here, we give a detailed characterization of the covering number in terms of the spectrum using the operator $T_m$.
Accordingly, we give the complexity of the set of 
%$f$'s
functions
satisfying the Convolution Assumption (Assumption \ref{ass:convolution}).
We extend the domain and the range of the operator $T_m$ to the whole space of $\LPi$, and define its power $T_m^{\beta}:\LPi \to \LPi$ for $\beta \in [0,1]$ as 
\begin{equation*}
T_m^\beta f := \sum_{k=1}^{\infty} \mu_{k,m}^\beta \langle f, \phi_{k,m}\rangle_{\LPi}  \phi_{k,m},~~~(f\in \LPi).
\end{equation*}
Moreover, we define a Hilbert space $\calH_{m,\beta}$ as 
$$
\textstyle \calH_{m,\beta} := \{\sum_{k=1}^{\infty} b_k \phi_{k,m} \mid \sum_{k=1}^\infty  \mu_{k,m}^{-\beta} b_k^2 \leq \infty \},
$$
and equip this space with the Hilbert space norm
$
\textstyle \left\|\sum_{k=1}^{\infty} b_k \phi_{k,m} \right\|_{\calH_{m,\beta}} := \sqrt{\sum_{k=1}^\infty  \mu_{k,m}^{-\beta} b_k^2}.
$
One can check that $\calH_{m,1} = \calH_m$. 
Here we define, for $R>0$,
\begin{equation}
\label{eq:defHmqR}
\calH_{m}^{q}(R) := \{f_m = T_m^{\frac{q}{2}}g_m \mid g_m\in \calH_m,~\|g_m\hnorm{m} \leq R \}.
\end{equation}
Then we obtain the following lemma.
\begin{Lemma}
\label{lemm:HmqEquiv}
$\calH_{m}^{q}(1)$ is equivalent to the unit ball of $\calH_{m,1+q}$: $\calH_{m}^{q}(1) = \{ f_m \in \calH_{m,1+q} \mid \|f_m \hnorm{m} \leq 1 \}.$
\end{Lemma}
This can be shown as follows. For all $f_m \in \calH_{m}^{q}(1)$, there exists $g_m \in \calH_m$ such that $f_m = T_m^{\frac{q}{2}}g_m$ and $\|g_m \hnorm{m}\leq 1$. 
Thus,
$g_m = (T_m^{\frac{q}{2}})^{-1} f_m = \sum_{k=1}^{\infty} \mu_{k,m}^{-\frac{q}{2}} \langle f, \phi_{k,m}\rangle_{\LPi}\phi_{k,m}$ and 
$1 \geq \|g_m\hnorm{m} = 
 \sum_{k=1}^{\infty} \mu_{k,m}^{-1} \langle g, \phi_{k,m}\rangle_{\LPi}^2 = \sum_{k=1}^{\infty} \mu_{k,m}^{-(1+q)} \langle f, \phi_{k,m}\rangle_{\LPi}^2$.
Therefore, $f\in \calH_m$ is in $\calH_{m}^{q}(1)$ if and only if the norm of $f$ in $\calH_{m,1+q}$ is well-defined and not greater than 1.

Now Theorem 15 of \cite{COLT:Steinwart+etal:2009} gives an upper bound of the covering number of the unit ball $\mathcal{B}_{\calH_{m,\beta}}$ in $\calH_{m,\beta}$ as 
$
\calN(\varepsilon,\mathcal{B}_{\calH_{m,\beta}},\LPi) \leq C \varepsilon^{-2 \frac{s}{\beta}},
$
where $C$ is a constant depending on $c,s,\beta$.
This inequality with $\beta = 1$ corresponds to \Eqref{eq:coveringcondition}.
Moreover, substituting $\beta=1+q$ into the above equation,
we have
%  Lemma \ref{lemm:HmqEquiv} gives
\begin{align}
\label{eq:coveringconditionHq}
\calN(\varepsilon,\calH_{m}^{q}(1),\LPi) \leq C \varepsilon^{-2 \frac{s}{1+q}}.
\end{align}
%This inequality is utilized to show the minimax optimal rate. 

\section{Proof of Lemma \ref{lem:incoherenceIneq}}
\label{sec:appendixLemm}
\begin{proof} {\bf (Lemma \ref{lem:incoherenceIneq})}
For $J = I^c$, we have
\begin{align}
P f^2  &=\|f_I \|_{\LPi}^2 + 2 \langle f_I , f_J \rangle_{\LPi} 
+ \|f_J\|_{\LPi}^2  %\notag \\ & 
\geq 
\|f_I  \|_{\LPi}^2 - 2 \rho(I) \| f_I\|_{\LPi}  \| f_J \|_{\LPi} 
+ \|f_J \|_{\LPi}^2 \notag \\
& \geq 
(1- \rho(I)^2) \| f_I \|_{\LPi}^2
\geq 
(1- \rho(I)^2) \kmin(I) \left(\sum_{m\in I}\| f_m \|_{\LPi}^2\right),
\label{eq:firstboundforbasic}
\end{align}
where we used the inequality of arithmetic and geometric mean in the second inequality.
%  but one. 
\end{proof}

\section{Talagrand's Concentration Inequality} % and Maximal Inequality}

\begin{Proposition}{\rm \bf (Talagrand's Concentration Inequality \citep{Talagrand2,BousquetBenett})}
\label{prop:TalagrandConcent}
Let $\calG$ be a function class on $\calX$ that is separable with respect to $\infty$-norm, and 
$\{x_i\}_{i=1}^n$ be i.i.d. random variables with values in $\calX$.
Furthermore, let $B\geq 0$ and $U\geq 0$ be 
$B := \sup_{g \in \calG} \EE[(g-\EE[g])^2]$ and $U := \sup_{g \in \calG} \|g\|_{\infty}$,
then there exists a universal constant $K$ such that, for $Z := \sup_{g\in \calG}\left|\frac{1}{n} \sum_{i=1}^n g(x_i) - \EE[g] \right|$, we have
\begin{align*}
P\left( Z \geq K\left[\EE[Z] + \sqrt{\frac{B t}{n}} + \frac{U t}{n} \right] \right) \leq e^{-t},
\end{align*}
for all $t > 0$.
\end{Proposition}

%\begin{Proposition}{\rm \bf (Maximal Inequality)}
%\label{prop:Maximal}
%There exists a universal constant $K$ such that for any finite class $\calF$ of bounded, measurable, squared-integrable functions, with $|\calF| \geq 2$ elements, 
%\begin{align}
%\EE\left[\sup_{f \in \calF} \left|\frac{1}{n} \sum_{i=1}^n f(x_i)\right|\right] \leq K 
%\left[\max_{f\in \calF} \frac{\|f\|_{\infty}}{n} \log(|\calF|) + \max_{f\in \calF} \|f\|_{L_2(P)} \sqrt{\frac{\log(|F|)}{n}}\right].
%\end{align}
%\end{Proposition}

\section{Proof of Theorem \ref{eq:TheConvergenceRateMain}}
\label{proof:TheConvergenceRateMain}

For a Hilbert space $\calG \subset L_2(P)$, let the {\it $i$-th entropy number} $e_i(\calG \to L(P))$ be the infimum of $\epsilon > 0$ for which 
$
\calN(\epsilon,\calB_{\calG},L_2(P)) \leq 2^{i-1},
$
where $\calB_{\calG}$ is the unit ball of $\calG$.
One can check that if the spectral assumption (A\ref{ass:specass}) holds, 
the $i$-th entropy number is bounded as 
\begin{align}
\label{eq:entropycondition}
e_i(\calH_m \to \LPi) \leq \tilde{c} i^{- \frac{1}{2s}}.
\end{align}
where $\tilde{c}$ is a constant depends on $s$ and $c$.

The following proposition is the key of the localization. 
\begin{Proposition}
\label{prop:localRadeBound}
Let $\calB_{\sigma,a,b} \subset \calH_m$ be a set such that $\calB_{\sigma,a,b} = \{ f_m \in \calH_m \mid \|f_m\|_{\LPi}\leq \sigma, \|f_m\hnorm{m} \leq a, \|f_m\|_{\infty} \leq b\} $.
Assume the Spectral Assumption (A\ref{ass:specass}), 
%there exist constants $0<s<1$ and $0 < \tilde{c}_s$ such that 
%\begin{align*}
%\EE_{D_n}[ e_i(\calH_m \to L_2(D_n))] \leq \tilde{c}_s i^{-\frac{1}{2s}}.
%\end{align*}
then there exist constants $\tilde{c}_s,C_s'$ depending only $s$ and $c$ such that 
\begin{align*}
\EE\left[\sup_{f_m \in \calB_{\sigma,a,b}} \left| \frac{1}{n}\sum_{i=1}^n \sigma_i f_m(x_i) \right|\right] \leq 
C_s' \left( \frac{ \sigma^{1-s} (\tilde{c}_s a)^s}{\sqrt{n}} \vee (\tilde{c}_s a)^{\frac{2s}{1+s}} b^{\frac{1-s}{1+s}} n^{-\frac{1}{1+s}} \right). 
\end{align*}
\end{Proposition}
\begin{proof} {\bf (Proposition \ref{prop:localRadeBound})}
Let $D_n$ be the empirical distribution: $D_n = \frac{1}{n}\sum_{i=1}^n \delta_{x_i}$.
To bound empirical processes, a bound of the entropy number with respect to the empirical $L_2$-norm is needed.
Corollary 7.31 of \cite{Book:Steinwart:2008} gives the following upper bound: 
%\begin{Proposition}
%\label{prop:upperboundofe}
%If there exists constants $0<s<1$ and $c \geq 1$ such that 
%$e_i(\calH_m \to \LPi) \leq c i^{- \frac{1}{2s}}$, then  
under the condition \eqref{eq:entropycondition}, there exists a constant $c_s > 0$ only depending on $s$ such that 
%\begin{align*}
%\EE_{D_n \sim \Pi^n}[e_i(\calH_m \to L_2(D_n))] \leq c_s \tilde{c} (\min(i,n))^{\frac{1}{2s}} i^{-\frac{1}{s}},
%\end{align*}
%in particular 
$$
\EE_{D_n \sim \Pi^n}[e_i(\calH_m \to L_2(D_n))] \leq c_s \tilde{c}  i^{-\frac{1}{2s}}.
$$
Finally this and Theorem 7.16 of \cite{Book:Steinwart:2008} gives the assertion. 
\end{proof}

Using the above proposition and the {\it peeling device}, we obtain the following lemma (see also \citet{AS:Meier+Geer+Buhlmann:2009}). 
\begin{Lemma}
\label{lemm:simpleRationBound}
Under the Spectral Assumption (Assumption \ref{ass:specass}), there exists a constant $C_s$ depending only on $s$ and $C$ such that for all $\lambdatmp > 0$
\begin{align*}
\EE\left[\sup_{f_m \in \calH_m: \|f_m \hnorm{m}\leq 1} \frac{|\frac{1}{n}\sum_{i=1}^n \sigma_i f_m(x_i)|}{\sqrt{\|f_m\|_{\LPi}^2 + \lambdatmp} } \right] 
\leq C_s \left(\frac{\lambdatmp^{-\frac{s}{2}}}{\sqrt{n}} \vee \frac{1}{\lambdatmp^{\frac{1}{2}} n^{\frac{1}{1+s}}}\right) .
\end{align*}
\end{Lemma}

\begin{proof} {\bf (Lemma \ref{lemm:simpleRationBound})} 
Let $\calH_m(\sigma) := \{ f_m \in \calH_m \mid \|f_m\hnorm{m} \leq 1, \|f_m\|_{\LPi} \leq \sigma \}$ and $z = 2^{1/s}>1$.
%We write $\tilde{c}_s = c_s \tilde{c}$ where $c_s$ and $\tilde{c}$ are the constants appeared in the proof of Proposition \ref{prop:localRadeBound}.
Then by noticing $\|f_m\|_{\infty} \leq \|f_m\hnorm{m}$, Proposition \ref{prop:localRadeBound} gives
\begin{align*}
&\EE\left[\sup_{f_m \in \calH_m: \|f_m \hnorm{m}\leq 1} \frac{|\frac{1}{n}\sum_{i=1}^n \sigma_i f_m(x_i)|}{\sqrt{\|f_m\|_{\LPi}^2 + \lambdatmp} } \right]  \notag \\
\leq &
\EE\left[\sup_{f_m \in \calH_m(\lambdatmp^{1/2})} \frac{|\frac{1}{n}\sum_{i=1}^n \sigma_i f_m(x_i)|}{\sqrt{\|f_m\|_{\LPi}^2 + \lambdatmp} } \right] 
%\notag \\ &
+\sum_{k=1}^{\infty} \EE\left[\sup_{f_m \in \calH_m(z^k \lambdatmp^{1/2}) \backslash \calH_m(z^{k-1} \lambdatmp^{1/2})} \frac{|\frac{1}{n}\sum_{i=1}^n \sigma_i f_m(x_i)|}{\sqrt{\|f_m\|_{\LPi}^2 + \lambdatmp} } \right] \notag \\
\leq &
C_s' \left( \frac{ \lambdatmp^{\frac{1-s}{2}} \tilde{c}_s^s}{\lambdatmp^{\frac{1}{2}} \sqrt{n}} \vee \frac{\tilde{c}_s^{\frac{2s}{1+s}} }{n^{\frac{1}{1+s}} \lambdatmp^{\frac{1}{2}}} \right)
%\notag \\ &
+\sum_{k=0}^{\infty} C_s' \left( \frac{ z^{k(1-s)} \lambdatmp^{\frac{1-s}{2}} \tilde{c}_s^s}{ \sqrt{n} z^k\lambdatmp^{\frac{1}{2}}} \vee \frac{\tilde{c}_s ^{\frac{2s}{1+s}} }{n^{\frac{1}{1+s}} z^k\lambdatmp^{\frac{1}{2}}} \right)  \notag \\
= &
C_s' \left( \tilde{c}_s^s \sqrt{\frac{ \lambdatmp^{-s} }{n}} \vee \tilde{c}_s^{\frac{2s}{1+s}} \left( \frac{\lambdatmp^{-\frac{1}{2}} }{n^{\frac{1}{1+s}}} \right)  \right)
%\notag \\ &
+\sum_{k=0}^{\infty} 
C_s' \left( \tilde{c}_s^s z^{-sk} \sqrt{\frac{ \lambdatmp^{-s} }{n}} \vee \tilde{c}_s^{\frac{2s}{1+s}} z^{-k} \left( \frac{\lambdatmp^{-\frac{1}{2}} }{n^{\frac{1}{1+s}}} \right)  \right) \notag \\
\leq &
2 C_s' \left( \frac{1}{1-z^{-s}}\tilde{c}_s^s \sqrt{\frac{ \lambdatmp^{-s} }{n}} +  \frac{1}{1-z^{-1}} \tilde{c}_s^{\frac{2s}{1+s}} \left( \frac{\lambdatmp^{-\frac{1}{2}} }{n^{\frac{1}{1+s}}} \right)  \right)
= 2 C_s' \left( 2 \tilde{c}_s^s \sqrt{\frac{ \lambdatmp^{-s} }{n}} +  \frac{2^{1/s}}{2^{1/s}-1} \tilde{c}_s^{\frac{2s}{1+s}} \left( \frac{\lambdatmp^{-\frac{1}{2}} }{n^{\frac{1}{1+s}}} \right)  \right) \notag \\
\leq &
 2 C_s' \left(2 \tilde{c}_s^s+\frac{2^{1/s}}{2^{1/s}-1} \tilde{c}_s^{\frac{2s}{1+s}}\right) \left(  
\sqrt{\frac{ \lambdatmp^{-s} }{n}} \vee  \left( \frac{\lambdatmp^{-\frac{1}{2}} }{n^{\frac{1}{1+s}}} \right)  \right).
\end{align*}
By setting $C_s \leftarrow 2 C_s' \left(2 \tilde{c}_s^s+\frac{2^{1/s}}{2^{1/s}-1} \tilde{c}_s^{\frac{2s}{1+s}}\right)$, we obtain the assertion.
\end{proof}

The above lemma immediately gives the following corollary.
\begin{Corollary}
Under the Spectral Assumption (Assumption \ref{ass:specass}), 
%There exists a constant $C_s$ that depends only $s$ and $c$ such that 
for all $\lambdatmp > 0$
\begin{align*}
\EE\left[\sup_{f_m \in \calH_m} \frac{|\frac{1}{n}\sum_{i=1}^n \sigma_i f_m(x_i)|}{\sqrt{\|f_m\|_{\LPi}^2 + \lambdatmp \|f_m\hnorm{m}^2} } \right] 
\leq C_s \left(\frac{\lambdatmp^{-\frac{s}{2}}}{\sqrt{n}} \vee \frac{1}{\lambdatmp^{\frac{1}{2}} n^{\frac{1}{1+s}}}\right),
\end{align*}
where $C_s$ is the constant appeared in the statement of Lemma \ref{lemm:simpleRationBound}, and we employed a convention such that $\frac{0}{0} = 0$.
\end{Corollary}

Moreover we obtain the following corollary. %The proof is given in Appendix \ref{sec:appendixMainTheorems}.
\begin{Corollary}
\label{eq:basicupperboundofexp}
Under the Spectral Assumption (Assumption \ref{ass:specass}), %There exists a constant $C_s$ that depends only $s$ and $c$ such that 
for all $\lambdatmp > 0$
\begin{align*}
\EE\left[\sup_{f_m \in \calH_m} \frac{|\frac{1}{n}\sum_{i=1}^n \epsilon_i f_m(x_i)|}{\sqrt{\|f_m\|_{\LPi}^2 + \lambdatmp \|f_m\hnorm{m}^2} } \right] 
\leq 2 C_s L  \left(\frac{\lambdatmp^{-\frac{s}{2}}}{\sqrt{n}} \vee \frac{1}{\lambdatmp^{\frac{1}{2}} n^{\frac{1}{1+s}}}\right),
\end{align*}
where $C_s$ is the constant appeared in the statement of Lemma \ref{lemm:simpleRationBound}.
\end{Corollary}
\begin{proof} {\bf(Corollary \ref{eq:basicupperboundofexp})}
Here we write $Pf = \EE[f]$ and $P_n f = \frac{1}{n} \sum_{i=1}^n f(x_i,y_i)$ for a function $f$.
Notice that $P \epsilon f_m= 0$, thus $\frac{1}{n}\sum_{i=1}^n \epsilon_i f_m(x_i) = (P_n - P)(\epsilon f_m)$.
By the symmetrization argument \cite[Lemma 2.3.1]{Book:VanDerVaart:WeakConvergence} and the contraction inequality \cite[Theorem 4.12]{Book:Ledoux+Talagrand:1991}, we obtain
\begin{align*}
\EE\left[\sup_{f_m \in \calH_m} \frac{|(P-P_n)( \epsilon f_m)|}{\sqrt{\|f_m\|_{\LPi}^2 + \lambdatmp \|f_m\hnorm{m}^2} } \right] 
=&
\EE\left[\sup_{f_m \in \calH_m} \left|(P-P_n) \frac{ \epsilon f_m}{\sqrt{\|f_m\|_{\LPi}^2 + \lambdatmp \|f_m\hnorm{m}^2} } \right| \right]  \\
\leq &
2 \EE\left[\sup_{f_m \in \calH_m} \left| \frac{ \frac{1}{n}\sum_{n=1}^n \sigma_i \epsilon_if_m(x_i)}{\sqrt{\|f_m\|_{\LPi}^2 + \lambdatmp \|f_m\hnorm{m}^2} } \right| \right] \\
\leq &
2 L \EE\left[\sup_{f_m \in \calH_m} \left| \frac{ \frac{1}{n}\sum_{n=1}^n \sigma_i f_m(x_i)}{\sqrt{\|f_m\|_{\LPi}^2 + \lambdatmp \|f_m\hnorm{m}^2} } \right| \right] \\
\leq &
2 C_s L  \left(\frac{\lambdatmp^{-\frac{s}{2}}}{\sqrt{n}} \vee \frac{1}{\lambdatmp^{\frac{1}{2}} n^{\frac{1}{1+s}}}\right).
\end{align*}
This gives the assertion. 
\end{proof}

From now on, we refer to $C_s$ as the constant appeared in the statement of Lemma \ref{lemm:simpleRationBound}.
We define $\tilde{\phi}_s$ as 
$$
%\tilde{\phi}_s = K\left[L (2 C_s  + (C_1 + 1)K) + 2 L\right].
\tilde{\phi}_s = 2KL(C_s  + 1 + C_1).
$$
Remind the definition of $\xi_n$ (\Eqref{eq:definitionxin}), then we obtain the following theorem. 
%The proof is given in Appendix \ref{sec:appendixMainTheorems}.

\begin{Theorem}
\label{eq:ffdiscrepancyBoundLone}
Under the Basic Assumption, the Spectral Assumption and the Supnorm Assumption,
%there exists a constant  
%$\tilde{\phi}_s$ that depends only $s,c,L,C_1$ such that  
when %$\lambdatmp = n^{-\frac{1}{1+q+s}} \vee (\frac{log(M)}{n})^{\frac{1}{1+q}}$ and 
$\frac{\log(M)}{\sqrt{n}} \leq 1$, 
we have 
for all $\lambdatmp>0$ and all $t \geq 1$
\begin{align*} 
&\left| \frac{1}{n}\sum_{i=1}^n \epsilon_i (\fhat_m(x_i) - \fstar_m(x_i)) \right|  
\leq 
\tilde{\phi}_{s}
 \xi_n(\lambdatmp) \sqrt{\|f_m\|_{\LPi}^2 + \lambdatmp \|f_m\hnorm{m}^2}
\max\left(1,\sqrt{t}, t/\sqrt{n}\right), \\
&(\forall f_m \in \calH_m, \forall m=1,\dots,M),
\end{align*}
with probability $1- \exp( - t)$. %$\exp(-n + \log(M))$.
Moreover we also have 
\begin{align}
&\EE\left[\max_m \sup_{f_m \in \calH_m} \frac{|\frac{1}{n}\sum_{i=1}^n \epsilon_i f_m(x_i)|}{\sqrt{\|f_m\|_{\LPi}^2 + \lambdatmp \|f_m\hnorm{m}^2} }  \right]
\leq 4 \tilde{\phi}_s \xi_n.
\end{align} 

\end{Theorem}
\begin{proof} {\bf (Theorem \ref{eq:ffdiscrepancyBoundLone})}
Since 
\begin{align}
& \frac{\| f_m \|_{\LPi}}{\sqrt{\|f_m\|_{\LPi}^2 + \lambdatmp \|f_m\hnorm{m}^2} } \leq 1, \\
& \frac{\| f_m \|_{\infty}}{\sqrt{\|f_m\|_{\LPi}^2 + \lambdatmp \|f_m\hnorm{m}^2} } \leq \frac{C_1 \| f_m \|_{\LPi}^{1-s}\| f_m \hnorm{m}^s}{\sqrt{\|f_m\|_{\LPi}^2 + \lambdatmp \|f_m\hnorm{m}^2} }
\mathop{\leq}^{\text{Young}} \frac{C_1 \lambdatmp^{-\frac{s}{2}} \sqrt{ \| f_m \|_{\LPi}^2 + \lambdatmp \| f_m \hnorm{m}^2}}{\sqrt{\|f_m\|_{\LPi}^2 + \lambdatmp \|f_m\hnorm{m}^2} } \notag \\
&\leq C_1 \lambdatmp^{-\frac{s}{2}},
\end{align}
applying Talagrand's concentration inequality (Proposition \ref{prop:TalagrandConcent}), we obtain 
\begin{align*}
&P\left( \sup_{f_m \in \calH_m} \frac{|\frac{1}{n}\sum_{i=1}^n \epsilon_i f_m(x_i)|}{\sqrt{\|f_m\|_{\LPi}^2 + \lambdatmp \|f_m\hnorm{m}^2} }   
\geq  K\left[2 C_s L \xi_n
+ \sqrt{\frac{L^2 t}{n}} + \frac{C_1 L  \lambdatmp^{-\frac{s}{2}} t}{n}   \right] \right) \notag \\ 
&\leq e^{-t}.
\end{align*} 
Therefore the uniform bound over all $m=1,\dots,M$ is given as 
\begin{align*}
&P\left( \max_m \sup_{f_m \in \calH_m} \frac{|\frac{1}{n}\sum_{i=1}^n \epsilon_i f_m(x_i)|}{\sqrt{\|f_m\|_{\LPi}^2 + \lambdatmp \|f_m\hnorm{m}^2} }   
\geq  K\left[2 C_s L \xi_n
+ \sqrt{\frac{L^2 t}{n}} + \frac{C_1 L  \lambdatmp^{-\frac{s}{2}} t}{n}   \right] \right) \notag \\ 
\leq & \sum_{m=1}^M P\left( \sup_{f_m \in \calH_m} \frac{|\frac{1}{n}\sum_{i=1}^n \epsilon_i f_m(x_i)|}{\sqrt{\|f_m\|_{\LPi}^2 + \lambdatmp \|f_m\hnorm{m}^2} }   
\geq  K\left[2 C_s L \xi_n
+ \sqrt{\frac{L^2 t}{n}} + \frac{C_1 L  \lambdatmp^{-\frac{s}{2}} t}{n}   \right] \right) \notag \\ 
\leq & M e^{-t}.
\end{align*} 
Setting $t\leftarrow t + \log(M)$, we have 
\begin{align}
&P\left( \max_m \sup_{f_m \in \calH_m} \frac{|\frac{1}{n}\sum_{i=1}^n \epsilon_i f_m(x_i)|}{\sqrt{\|f_m\|_{\LPi}^2 + \lambdatmp \|f_m\hnorm{m}^2} }   
\geq  K\left[2 C_s L \xi_n
+ \sqrt{\frac{L^2 (t+\log(M))}{n}} + \frac{C_1 L  \lambdatmp^{-\frac{s}{2}} (t+\log(M))}{n}   \right] \right) \notag \\
& \leq e^{-t}. 
\label{eq:maxineqtmp}
\end{align} 
Now 
\begin{align*}
\sqrt{\frac{L^2 (t+\log(M))}{n}} + \frac{C_1 L  \lambdatmp^{-\frac{s}{2}} (t+\log(M))}{n} 
&\leq
L\sqrt{\frac{t}{n}} + L \sqrt{\frac{\log(M)}{n}} + \frac{C_1 L  \lambdatmp^{-\frac{s}{2}}}{\sqrt{n}}\left( \frac{t}{\sqrt{n}} + \frac{\log(M)}{\sqrt{n}}  \right) \\
&\leq
\xi_n \left( L  \sqrt{t} + L + C_1 L  \frac{t}{\sqrt{n}} +   C_1 L \right)
\leq 
\xi_n \left( 2L + 2C_1 L \right)\eta(t).
\end{align*}
where we used $\frac{\log(M))}{\sqrt{n}} \leq 1$ in the second inequality.
Thus \Eqref{eq:maxineqtmp} implies 
\begin{align*}
&P\left( \max_m \sup_{f_m \in \calH_m} \frac{|\frac{1}{n}\sum_{i=1}^n \epsilon_i f_m(x_i)|}{\sqrt{\|f_m\|_{\LPi}^2 + \lambdatmp \|f_m\hnorm{m}^2} }   
\geq  K(2 C_s L + 2L + 2C_1 L)\xi_n \eta(t) \right) \notag \\
& \leq e^{-t}. 
\end{align*} 
By substituting $\tilde{\phi}_s = 2KL(C_s  + 1 + C_1)$, we obtain 
\begin{align}
&P\left( \max_m \sup_{f_m \in \calH_m} \frac{|\frac{1}{n}\sum_{i=1}^n \epsilon_i f_m(x_i)|}{\sqrt{\|f_m\|_{\LPi}^2 + \lambdatmp \|f_m\hnorm{m}^2} }   
\geq  \tilde{\phi}_s \xi_n \eta(t) \right) \leq e^{-t},
\label{eq:maxineqtmp2}
\end{align} 
which gives the first assertion. 

Next we show the second assertion.
\Eqref{eq:maxineqtmp2} implies that  
\begin{align}
&\EE\left[\max_m \sup_{f_m \in \calH_m} \frac{|\frac{1}{n}\sum_{i=1}^n \epsilon_i f_m(x_i)|}{\sqrt{\|f_m\|_{\LPi}^2 + \lambdatmp \|f_m\hnorm{m}^2} }  \right]
%&
%\leq 
%\tilde{\phi}_s \xi_n  + \int_{0}^{\infty} \tilde{\phi}_s \xi_n \eta(t) e^{-t} \dd t \\
\leq 
\tilde{\phi}_s \xi_n  + 
\sum_{t=0}^{\infty} e^{-t} \tilde{\phi}_s \xi_n \eta(t+1) \\
&
\leq 
\tilde{\phi}_s \xi_n  + 
\tilde{\phi}_s \xi_n \sum_{t=0}^{\infty} e^{-t}  (t+1) 
%\int_0^{\infty} e^{-t+1} \tilde{\phi}_s \xi_n \eta(t+2) \dd t 
\leq 
4 \tilde{\phi}_s \xi_n,
%\int_{0}^{\infty} \tilde{\phi}_s \xi_n \eta(t) e^{-t} \dd t.
\end{align} 
where we used $\eta(t+1) = \max\{1,\sqrt{t+1},(t+1)/\sqrt{n}\}\leq t+1$ in the second inequality.
Thus we obtain the assertion.
\end{proof}

Moreover we obtain the following bound for the difference of the empirical and the expectation $L_2$-norm.
%The proof is given in Appendix \ref{sec:appendixMainTheorems}.
Let $\tilde{\phi}_s'$ be
$$
\tilde{\phi}_s' = K\left[16 KC_1(C_s  + 1 + C_1)    +  C_1 + C_1^2  \right].
%K\left[8 C_1 \tilde{\phi}_s   +  C_1 + C_1^2  \right].
$$
We define $\zeta_n(r,\lambdatmp)$ as
\begin{align*}
\zeta_n(r,\lambdatmp) := \min\left( \frac{r^2 \log(M)}{n\xi_n(\lambdatmp)^4\tilde{\phi}_s'^2}, \frac{r}{\xi_n(\lambdatmp)^2\tilde{\phi}_s'}\right).
\end{align*}

\begin{Theorem}
\label{eq:fsquareBounds}
Under the Spectral Assumption and the Supnorm Assumption, 
%there exists a constant $\tilde{\phi}_s'$ that depends only $s,c,K,C_1$ such that 
when $\frac{\log(M)}{\sqrt{n}} \leq 1$, %$\lambdatmp = n^{-\frac{1}{1+q+s}} \vee (\frac{\log(M)}{n})^{\frac{1}{1+q}}$ and $\frac{\log(M)}{n} \leq 1$,
for all $\lambdatmp > 0$ we have 
\begin{align*}
&\left| \textstyle \left\|\sum_{m=1}^M f_m \right\|_n^2 - \left\|\sum_{m=1}^M f_m \right\|_{\LPi}^2 \right|  \leq 
\max(\tilde{\phi}_s' \sqrt{n}\xi_n^2(\lambdatmp),r) \left(
 \sum_{m=1}^M \sqrt{ \|f_m\|_{\LPi}^2 + \lambdatmp \|f_m\hnorm{m}^2}\right)^2,~~\\
&(\forall f_m \in \calH_m~(m=1,\dots,M)),
\end{align*}
with probability $1- \exp( - \zeta_n(r,\lambdatmp) )$. %$\exp(-n + \log(M))$.
\end{Theorem}
\begin{proof} {\bf (Theorem \ref{eq:fsquareBounds})}
\begin{align}
&\EE\left[\sup_{f_m \in \calH_m} \frac{\left| \textstyle \left\|\sum_{m=1}^M f_m \right\|_n^2 - \left\|\sum_{m=1}^M f_m \right\|_{\LPi}^2 \right| }{
\left(\sum_{m=1}^M \sqrt{\|f_m\|_{\LPi}^2 + \lambdatmp \|f_m\hnorm{m}^2} \right)^2}  \right] \notag \\
\leq &  
2 \EE\left[ \sup_{f_m \in \calH_m} \frac{\textstyle  \left| \frac{1}{n} \sum_{i=1}^n \sigma_i ( \sum_{m=1}^M f_m(x_i))^2   \right| }{
\left(\sum_{m=1}^M \sqrt{\|f_m\|_{\LPi}^2 + \lambdatmp \|f_m\hnorm{m}^2} \right)^2}  \right] 
\notag \\
\leq &
\sup_{f_m \in \calH_m} 
\frac{\textstyle  \left\| \sum_{m=1}^M f_m \right\|_{\infty} }{
\sum_{m=1}^M \sqrt{\|f_m\|_{\LPi}^2 + \lambdatmp \|f_m\hnorm{m}^2} } \times
2 \EE\left[ 
\sup_{f_m \in \calH_m} 
\frac{\textstyle  \left| \frac{1}{n} \sum_{i=1}^n \sigma_i ( \sum_{m=1}^M f_m(x_i))   \right| }{
\sum_{m=1}^M \sqrt{\|f_m\|_{\LPi}^2 + \lambdatmp \|f_m\hnorm{m}^2} }  \right],
\label{eq:squareUpperBoundtmp}
\end{align} 
where we used the contraction inequality in the last line  \cite[Theorem 4.12]{Book:Ledoux+Talagrand:1991}.
Here we notice that 
\begin{align*}
\left\| \sum_{m=1}^M f_m \right\|_{\infty} &\leq \sum_{m=1}^M C_1 \|f_m\|_{\LPi}^{1-s} \|f_m\hnorm{m}^s
\leq \sum_{m=1}^M C_1 \lambdatmp^{-\frac{s}{2}} \sqrt{\|f_m\|_{\LPi}^{2(1-s)} (\lambdatmp \|f_m\hnorm{m}^{2})^s}  \\
&
\leq \sum_{m=1}^M C_1 \lambdatmp^{-\frac{s}{2}} \sqrt{(1-s) \|f_m\|_{\LPi}^{2} + s \lambdatmp \|f_m\hnorm{m}^{2}}  
\leq \sum_{m=1}^M C_1 \lambdatmp^{-\frac{s}{2}} \sqrt{ \|f_m\|_{\LPi}^{2} + \lambdatmp \|f_m\hnorm{m}^{2}},
\end{align*}
where we used Young's inequality $a^{1-s} b^s \leq (1-s) a + s b$ in the second line.
Thus the RHS of the inequality \eqref{eq:squareUpperBoundtmp} can be upper bounded by 
\begin{align*}
&2C_1 \lambdatmp^{-\frac{s}{2}} \EE\left[ \sup_{f_m \in \calH_m} 
\frac{\textstyle  \left| \frac{1}{n} \sum_{i=1}^n \sigma_i ( \sum_{m=1}^M f_m(x_i))   \right| }{
\sum_{m=1}^M \sqrt{\|f_m\|_{\LPi}^2 + \lambdatmp \|f_m\hnorm{m}^2} }  \right] \\
\leq & 
2 C_1 \lambdatmp^{-\frac{s}{2}} \EE\left[ \sup_{f_m \in \calH_m} \max_m 
\frac{\textstyle  \left| \frac{1}{n} \sum_{i=1}^n \sigma_i f_m(x_i)   \right| }{
\sqrt{\|f_m\|_{\LPi}^2 + \lambdatmp \|f_m\hnorm{m}^2} }  \right],
%\\ & C_s   \left(\frac{\lambdatmp^{-\frac{1+s}{2}}}{\sqrt{n}} \vee \frac{1}{\lambda n^{\frac{1}{1+s}}}\right).
\end{align*}
where we used the relation $\frac{\sum_{m} a_m}{\sum_m b_m} \leq \max_m(\frac{a_m}{b_m})$
for all $a_m \geq 0$ and $b_m \geq 0$ with a convention $\frac{0}{0}=0$.
Therefore, by $\frac{\log(M)}{\sqrt{n}} \leq 1$ and 
Theorem \ref{eq:ffdiscrepancyBoundLone} where $\sigma_i$ is substituted into $\epsilon_i$, the right hand side is upper bounded by 
$16 KC_1(C_s  + 1 + C_1)  \lambdatmp^{-\frac{s}{2}} \xi_n$. % 2C_1 (2 C_s  + (C_1 + 1)K) \lambdatmp^{\frac{1+q-s}{2}}$.
Here we again apply Talagrand's concentration inequality, then we have
\begin{align}
&P\Biggl(  \sup_{f_m \in \calH_m} \frac{ \left| \textstyle \left\|\sum_{m=1}^M f_m \right\|_n^2 - \left\|\sum_{m=1}^M f_m \right\|_{\LPi}^2 \right| }
{\left(\sum_{m=1}^M \sqrt{\|f_m\|_{\LPi}^2 + \lambdatmp \|f_m\hnorm{m}^2} \right)^2} \notag \\  
&~~~~~~~\geq K\left[16 KC_1(C_s  + 1 + C_1)  \lambdatmp^{-\frac{s}{2}}\xi_n  + \sqrt{\frac{t}{n}} C_1 \lambdatmp^{-\frac{s}{2}} + \frac{C_1^2 \lambdatmp^{-s} t}{n}   \right] \Biggr) \leq e^{-t},
\label{eq:squareTalagrands}
\end{align}
where we substituted the following upper bounds of $B$ and $U$:
\begin{align*}
B^2=&\sup_{f_m \in \calH_m} \EE \left[ \left( \frac{(\sum_{m=1}^M f_m)^2 }
{\left(\sum_{m=1}^M \sqrt{\|f_m\|_{\LPi}^2 + \lambdatmp \|f_m\hnorm{m}^2} \right)^2}\right)^2 \right] \\
\leq & 
\sup_{f_m \in \calH_m} \EE \left[ \frac{(\sum_{m=1}^M f_m)^2 } 
{\left( \sum_{m=1}^M  \|f_m\|_{\LPi} \right)^2} 
\frac{(\|\sum_{m=1}^M f_m\|_{\infty})^2 } 
{ \left( \sum_{m=1}^M \sqrt{ \|f_m\|_{\LPi}^2 + \lambdatmp \|f_m\hnorm{m}^2}\right)^2}  \right] \\
\leq & 
\sup_{f_m \in \calH_m}  \frac{\left(\sum_{m=1}^M \|f_m\|_{\LPi}\right)^2 } 
{\left( \sum_{m=1}^M  \|f_m\|_{\LPi} \right)^2} 
\frac{(\sum_{m=1}^M C_1 \lambdatmp^{-\frac{s}{2}} \sqrt{ \|f_m\|_{\LPi}^{2} + \lambdatmp \|f_m\hnorm{m}^{2}})^2 } 
{ \left( \sum_{m=1}^M \sqrt{ \|f_m\|_{\LPi}^2 + \lambdatmp \|f_m\hnorm{m}^2}\right)^2}  \\
\leq & C_1^2 \lambdatmp^{-s},
\end{align*}
where in the second inequality we used the relation $\EE[(\sum_{m=1}^M f_m)^2] = \EE[ \sum_{m,m'=1}^M f_m f_{m'}] \leq \sum_{m,m'=1}^M \|f_m\|_{\LPi} \|f_{m'}\|_{\LPi} = (\sum_{m=1}^M \|f_m\|_{\LPi})^2$,
and
\begin{align*}
U = &\sup_{f_m \in \calH_m} \left\| \frac{(\sum_{m=1}^M f_m)^2 }
{\left(\sum_{m=1}^M \sqrt{\|f_m\|_{\LPi}^2 + \lambdatmp \|f_m\hnorm{m}^2} \right)^2}\right\|  
\leq
\sup_{f_m \in \calH_m}  \frac{(\sum_{m=1}^M C_1 \lambdatmp^{-\frac{s}{2}} \sqrt{ \|f_m\|_{\LPi}^{2} + \lambdatmp \|f_m\hnorm{m}^{2}})^2 }
{\left(\sum_{m=1}^M \sqrt{\|f_m\|_{\LPi}^2 + \lambdatmp \|f_m\hnorm{m}^2} \right)^2}  \\	
\leq 
&
C_1^2 \lambdatmp^{-s}.
\end{align*}
Now notice that 
\begin{align*}
 & K\left[16 KC_1(C_s  + 1 + C_1)   \lambdatmp^{-\frac{s}{2}}\xi_n  + \sqrt{\frac{t}{n}} C_1 \lambdatmp^{-\frac{s}{2}} + \frac{C_1^2 \lambdatmp^{-s} t}{n} \right]\\
\leq & \sqrt{n} K\left[16 KC_1(C_s  + 1 + C_1)   \frac{\lambdatmp^{-\frac{s}{2}}}{\sqrt{n}} \xi_n  + \sqrt{\frac{t}{\log(M)}} C_1 \xi_n \sqrt{\frac{\log(M)}{n}}+ \frac{C_1^2 \xi_n^2 t}{\sqrt{n}}\right] \\
\leq & \sqrt{n} K\left[16 KC_1(C_s  + 1 + C_1)    + \sqrt{\frac{t}{\log(M)}} C_1 + \frac{C_1^2 t}{\sqrt{n}}\right]\xi_n^2.
\end{align*}
Therefore %for $t = n \lambdatmp^{(1+q)}$, %^{\frac{1+s}{2}}$, 
%the above inequality implies 
\Eqref{eq:squareTalagrands} gives the following inequality 
\begin{align*}
&\sup_{f_m \in \calH_m} \frac{ \left| \textstyle \left\|\sum_{m=1}^M f_m \right\|_n^2 - \left\|\sum_{m=1}^M f_m \right\|_{\LPi}^2 \right| }
{\left(\sum_{m=1}^M \sqrt{\|f_m\|_{\LPi}^2 + \lambdatmp \|f_m\hnorm{m}^2} \right)^2}  \notag \\
\leq &
K\left[16 KC_1(C_s  + 1 + C_1)     +  C_1 + C_1^2  \right] \sqrt{n} \xi_n^2 \max(1,\sqrt{t/\log(M)},t/\sqrt{n}). %\lambdatmp^{\frac{1+q-s}{2}}.
%K\left[L (2 C_s  + 2K) + 2 L\right]
\end{align*} 
with probability $1 - \exp( -t)$.
By substituting $\tilde{\phi}_s' = K\left[16 KC_1(C_s  + 1 + C_1)    +  C_1 + C_1^2  \right]$ and $t=\zeta_n(r,\lambdatmp)$, we have %the assertion. 
\begin{align*}
&\sup_{f_m \in \calH_m} \frac{ \left| \textstyle \left\|\sum_{m=1}^M f_m \right\|_n^2 - \left\|\sum_{m=1}^M f_m \right\|_{\LPi}^2 \right| }
{\left(\sum_{m=1}^M \sqrt{\|f_m\|_{\LPi}^2 + \lambdatmp \|f_m\hnorm{m}^2} \right)^2}  \notag \\
\leq &
\tilde{\phi}_s' \sqrt{n} \xi_n^2 \max\left(1,\sqrt{\frac{\zeta_n(r,\lambdatmp)}{\log(M)}},\frac{\zeta_n(r,\lambdatmp)}{\sqrt{n}}\right)
\leq
\tilde{\phi}_s' \sqrt{n} \xi_n^2 \max\left(1, \frac{r}{\tilde{\phi}_s' \sqrt{n} \xi_n^2}\right) \leq
\max\left(\tilde{\phi}_s' \sqrt{n} \xi_n^2 , r\right). %\lambdatmp^{\frac{1+q-s}{2}}.
%K\left[L (2 C_s  + 2K) + 2 L\right]
\end{align*} 
with probability $1 - \exp( -\zeta_n(r,\lambdatmp))$.

\end{proof}

Now we define 
\begin{align*}
%\phi_s := \max \left( KL\left[(2 C_s  + (C_1 + 1)K) + 2 \right] ,K\left[2C_1 (2 C_s  + (C_1 + 1)K)   +  C_1 + C_1^2  \right],1 \right).
\phi_s := \max \left( \tilde{\phi}_s',\tilde{\phi}_s,1\right)
= \max \left( K\left[16 KC_1(C_s  + 1 + C_1) +  C_1 + C_1^2\right], 2KL(C_s  + 1 + C_1), 1\right),
\end{align*}
where $K$ is the universal constant appeared in Talagrand's concentration inequality (Proposition \ref{prop:TalagrandConcent}).
%and the maximal inequality (Proposition \ref{prop:Maximal}).
We define events $\scrE_1(t)$ and $\scrE_2(r)$ as 
\begin{align*}
&
\scrE_1(t) = \left\{ 
\left| \frac{1}{n}\sum_{i=1}^n \epsilon_i f_m(x_i)  \right|  \leq 
\eta(t) \phi_{s}
\xi_n \sqrt{\|f_m\|_{\LPi}^2 + \lambdatmp \|f_m\hnorm{m}^2}, \forall f_m \in \calH_m,\forall m=1,\dots,M  \right\}, \\
&
\scrE_2(r) = \Bigg\{ 
\left| \textstyle \left\|\sum_{m=1}^M f_m \right\|_n^2 - \left\|\sum_{m=1}^M f_m \right\|_{\LPi}^2 \right| \leq 
\max( \phi_{s}\sqrt{n} \xi_n^2,r)
\left(\sum_{m=1}^M \sqrt{\|f_m\|_{\LPi}^2 + \lambdatmp \|f_m\hnorm{m}^2}\right)^2, \\
&~~~~~~~~~~~~~~~~~~ \forall f_m \in \calH_m,\forall m=1,\dots,M \Bigg\}.
\end{align*}
Theorems \ref{eq:ffdiscrepancyBoundLone} and \ref{eq:fsquareBounds} give that 
$P(\scrE_1(t)) \geq 1- e^{-t}$ and $P(\scrE_2(r)) \geq 1-\exp(-\zeta_n(r,\lambdatmp))$ under some conditions.

The next lemma gives a bound of irrelevant components ($m\in I_0^c$) of $\fhat$ in terms of the relevant components.
\begin{Lemma}
\label{lemm:basicTheoremNormBound}
Set $\lambdaone =4 \phi_s \eta(t) \xi_n(\lambdatmp)$, $\lambdatwo = \lambdatmp$, $\lambdathree = \lambdatmp$ for arbitrary $\lambdatmp >0$.
Then for all $n$ and $r(\geq 0)$ such that
$\frac{\log(M)}{\sqrt{n}} \leq 1$ and $\max(\phi_s \sqrt{n} \xi_n^2(\lambdatmp),r) \leq \frac{1}{2}$,
we have 
\begin{align}
& \sum_{m=1}^M \sqrt{\|\fstar_m - \fhat_m\|_{\LPi}^2 + \lambdatwo \|\fstar_m - \fhat_m\hnorm{m}^2}% + \sum_{m\in I_0} \lambdathree \|\fhat_m - \fstar_m\hnorm{m}^2
\notag \\
\leq 
&   
8 \sum_{m\in I_0} \left(1+ \frac{\lambdathree^{\frac{1+q}{2}}\|\gstar_m\hnorm{m}}{\lambdaone}\right) \left(\sqrt{\|\fstar_m - \fhat_m\|_{\LPi}^2 + \lambdatwo \|\fstar_m  - \fhat_m \hnorm{m}^2}\right),
\label{eq:ResidDominatedBound}
\end{align}
with probability $1 - \exp( - t) - \exp( - \zeta_n(r,\lambdatmp))$.
\end{Lemma}
\begin{proof} {\bf (Lemma \ref{lemm:basicTheoremNormBound})}
On the event $\scrE_2(r)$, for all $f_m \in \calH_m$ we obtain the upper bound of the regularization term as  
\begin{align}
&\sqrt{\|f_m\|_n^2 + \lambdatwo \|f_m\hnorm{m}^2} \\
&
\leq \sqrt{\|f_m\|_{\LPi}^2 + 
\max(\phi_s \sqrt{n} \xi_n^2(\lambdatmp),r)(\|f_m\|_{\LPi}^2 + \lambdatmp \|f_m\hnorm{m}^2)
+ \lambdatwo \|f_m\hnorm{m}^2} \notag \\
&
\leq \sqrt{ \frac{3}{2}(\|f_m\|_{\LPi}^2 + 
\lambdatwo \|\fhat_m\hnorm{m}^2)},
\label{eq:upperBoundOfLone}
\end{align}
because $\max(\phi_s \sqrt{n} \xi_n^2(\lambdatmp),r) \leq \frac{1}{2}$ and $\lambdatmp = \lambdatwo$.
On the other hand, we also obtain a lower bound as 
\begin{align}
&\sqrt{\|f_m\|_n^2 + \lambdatwo \|f_m\hnorm{m}^2} \\
&
\geq \sqrt{\|f_m\|_{\LPi}^2 -
\max(\phi_s \sqrt{n} \xi_n^2(\lambdatmp),r) (\|f_m\|_{\LPi}^2 + \lambdatmp \|f_m\hnorm{m}^2)
+ \lambdatwo \|f_m\hnorm{m}^2} \notag \\
&
\geq \sqrt{ \frac{1}{2}(\|f_m\|_{\LPi}^2 + 
\lambdatwo \|\fhat_m\hnorm{m}^2)},
\label{eq:lowerBoundOfLone}
\end{align}
for all $f_m \in \calH_m$.

Note that, since $\fhat$ minimizes the objective function,
\begin{align}
&\|\fhat - \fstar \|_n^2 + \sum_{m=1}^M( \lambdaone \sqrt{\|\fhat_m\|_n^2 + \lambdatwo \|\fhat_m\hnorm{m}^2} + \lambdathree  \|\fhat_m\hnorm{m}^2 ) \notag \\
\leq 
&\frac{1}{n}\sum_{n=1}^n \sum_{m=1}^M\epsilon_i (\fhat_m(x_i) - \fstar_m(x_i)) + 
 \sum_{m\in I_0} (\lambdaone \sqrt{\|\fstar_m\|_n^2 + \lambdatwo \|\fstar_m\hnorm{m}^2} + \lambdathree  \|\fstar_m\hnorm{m}^2).
\label{eq:basicineq}
\end{align}
This implies  %On the other hand, \Eqref{eq:basicineq} gives 
\begin{align*}
&\|\fhat - \fstar \|_n^2 + \sum_{m\in I_0^c}  \lambdaone \sqrt{\|\fhat_m\|_n^2 + \lambdatwo \|\fhat_m\hnorm{m}^2} \\
\leq 
&\frac{1}{n}\sum_{n=1}^n \sum_{m=1}^M\epsilon_i (\fhat_m(x_i) - \fstar_m(x_i))  \\
&+ \sum_{m\in I_0} (\lambdaone \sqrt{\|\fstar_m - \fhat_m\|_n^2 + \lambdatwo \|\fstar_m  - \fhat_m \hnorm{m}^2} + \lambdathree  (\|\fstar_m\hnorm{m}^2 - \|\fhat_m\hnorm{m}^2) ).
\end{align*}
Thus on the event $\calE_1(t)$ and $\calE_2(r)$, by \Eqref{eq:upperBoundOfLone} and \Eqref{eq:lowerBoundOfLone}, we have 
\begin{align}
&\|\fhat - \fstar \|_n^2 + \frac{1}{2} \sum_{m\in I_0^c}  \lambdaone \sqrt{\|\fhat_m\|_{\LPi}^2 + \lambdatwo \|\fhat_m\hnorm{m}^2} \notag \\
\leq 
&   \sum_{m=1}^M \eta(t) \phi_s \xi_n  \sqrt{\|\fhat_m \!- \fstar_m \|_{\LPi}^2 \!+ \lambdatwo \|\fhat_m - \fstar_m \hnorm{m}^2} + 
\notag
\\& \sum_{m\in I_0} \!(\frac{3}{2}\lambdaone \! \sqrt{\|\fstar_m \!- \fhat_m\|_{\LPi}^2 \!+\! \lambdatwo \|\fstar_m \! - \fhat_m \hnorm{m}^2} + \lambdathree  (2\langle \fstar_m, \fstar_m \!\! - \fhat_m \rangle_{\calH_m} 
\! \!\!- \|\fhat_m \! - \! \fstar_m\hnorm{m}^2) ) 
\label{eq:basicBoundpre}
\\
 \Rightarrow & \notag \\
&\frac{1}{4} \sum_{m\in I_0^c}  \lambdaone \sqrt{\|\fhat_m\|_{\LPi}^2 + \lambdatwo \|\fhat_m\hnorm{m}^2}% + \sum_{m\in I_0} \lambdathree \|\fhat_m - \fstar_m\hnorm{m}^2
\notag \\
\leq 
&   
 \sum_{m\in I_0} \left(\frac{7}{4}\lambdaone \sqrt{\|\fstar_m - \fhat_m\|_{\LPi}^2 + \lambdatwo \|\fstar_m  - \fhat_m \hnorm{m}^2} + 2 \lambdathree \langle T_m^{\frac{q}{2}}
\gstar_m, \fstar_m - \fhat_m \rangle_{\calH_m} \right).\notag
\end{align}
Now by the Young's inequality for positive symmetric operator, we have 
\begin{align}
\lambdathree^{1-q} T_m^{q} & = \lambdathree^{\frac{1}{2}} \left(\lambdathree^{-\frac{1}{2}} T_m
\lambdathree^{-\frac{1}{2}}\right)^q \lambdathree^{\frac{1}{2}} \notag \\
 & \preceq q T_m + (1-q) \lambdathree. \notag 
 \end{align}
Thus 
\begin{align}
&\lambdathree \langle 	
\fstar_m, \fstar_m - \fhat_m \rangle_{\calH_m}\notag \\
=& \lambdathree \langle T_m^{\frac{q}{2}}
\gstar_m, \fstar_m - \fhat_m \rangle_{\calH_m} \notag\\
 \leq 
& \lambdathree^{\frac{1+q}{2}} \|\gstar_m \hnorm{m}\| \lambdathree^{\frac{1-q}{2}} T_m^{\frac{q}{2}}(\fstar_m - \fhat_m) \hnorm{m} \notag\\
\leq 
&
\lambdathree^{\frac{1+q}{2}}\|\gstar_m \hnorm{m} \sqrt{\langle \fstar_m - \fhat_m,  \left(q T_m + (1-q) \lambdathree\right) \fstar_m - \fhat_m \rangle } \notag\\
=
&
\lambdathree^{\frac{1+q}{2}} \|\gstar_m \hnorm{m} \sqrt{ q \| \fstar_m - \fhat_m \|_{\LPi}^2 + (1-q) \lambdathree  \| \fstar_m - \fhat_m \hnorm{m}^2} \notag \\
\leq
&
\lambdathree^{\frac{1+q}{2}} \|\gstar_m \hnorm{m} \sqrt{ \| \fstar_m - \fhat_m \|_{\LPi}^2 + \lambdathree  \| \fstar_m - \fhat_m \hnorm{m}^2}.
\label{eq:gstarYoung}
\end{align}
Therefore we have 
\begin{align*}
&\frac{1}{4} \sum_{m\in I_0^c}  \lambdaone \sqrt{\|\fhat_m\|_{\LPi}^2 + \lambdatwo \|\fhat_m\hnorm{m}^2}% + \sum_{m\in I_0} \lambdathree \|\fhat_m - \fstar_m\hnorm{m}^2
\\
\leq 
&   
 \sum_{m\in I_0} \left(\frac{7}{4}\lambdaone + 2 \lambdathree^{\frac{1+q}{2}} \|\gstar_m \hnorm{m} \right) \sqrt{\|\fstar_m - \fhat_m\|_{\LPi}^2 + \lambdatwo \|\fstar_m  - \fhat_m \hnorm{m}^2}.
\end{align*}
with probability $1 - \exp( - t) - \exp( - \zeta_n(r,\lambdatmp))$.
The assertion is obvious from this bound.
\end{proof}

The next theorem immediately gives Theorem \ref{eq:TheConvergenceRateMain}.
\begin{Theorem}
\label{eq:TheConvergenceRateLambda}
Let $\lambdaone =4 \phi_s \eta(t) \xi_n(\lambdatmp)$, $\lambdatwo = \lambdatmp$, $\lambdathree = \lambdatmp$ for arbitrary $\lambdatmp > 0$. 
Then %re exists a constant $\widehat{C}_1$ depending $s,c,K,L,R,\kappa(I_0),\rho(I_0)$ such that 
for all $n$ and $r(\geq0)$ satisfying $\frac{\log(M)}{\sqrt{n}} \leq 1$ and the following inequality: %that satisfies  
\begin{align}
%\frac{\phi_s C_4^2}{(1-\rho(I_0))^2 \kappa(I_0)}d \sqrt{n} \xi_n^2 < \frac{1}{8},
\frac{128 \max(\phi_s \sqrt{n} \xi_n^2(\lambdatmp),r) \left( d + \frac{\lambdathree^{1+q}}{\lambdaone^2}\sum_{m=1}^M\|\gstar_m\hnorm{m}^2  \right)}{(1- \rho(I_0)^2) \kmin(I_0)} \leq \frac{1}{8}, 
\label{eq:ForLargen}
\end{align}
we have 
\begin{align}
\|\fhat - \fstar \|_{\LPi}^2 \leq \frac{48}{(1-\rho(I_0))^2 \kappa(I_0)} \left(d  \lambdaone^2 +  \lambdathree^{1+q}  \sum_{m=1}^M \|\gstar_m \hnorm{m}^2 \right), 
\label{eq:TheBoundlambda}
\end{align}
with probability $1- \exp(- t) - \exp(- \zeta_n(r,\lambdatmp))$ for all $t \geq 1$.
\end{Theorem}

\begin{proof} {\bf (Theorem \ref{eq:TheConvergenceRateLambda})}
By \Eqref{eq:basicBoundpre}, we have 
\begin{align*}
&\|\fhat - \fstar \|_{\LPi}^2 + \sum_{m \in I_0^c}( \lambdaone \sqrt{\|\fhat_m\|_n^2 + \lambdatwo \|\fhat_m\hnorm{m}^2} + \lambdathree  \|\fhat_m\hnorm{m}^2 ) 
+\sum_{m \in I_0} \lambdathree \|\fhat_m - \fstar_m \hnorm{m}^2 
\notag \\
\leq 
&( \|\fhat - \fstar \|_{\LPi}^2 - \|\fhat - \fstar \|_{n}^2 ) + 
\frac{1}{n}\sum_{n=1}^n \sum_{m=1}^M\epsilon_i (\fhat_m(x_i) - \fstar_m(x_i))  \notag \\ 
&+ \sum_{m\in I_0} (\lambdaone \sqrt{\|\fhat_m - \fstar_m\|_n^2 + \lambdatwo \|\fhat_m - \fstar_m\hnorm{m}^2} + \lambdathree  2 \langle \fstar_m,  \fstar_m - \fhat_m \rangle_{m}).
\end{align*}
Here on the event $\scrE_2(r)$, the above inequality gives 
\begin{align}
&\|\fhat - \fstar \|_{\LPi}^2 + \frac{1}{2} \sum_{m \in I_0^c}( \lambdaone \sqrt{\|\fhat_m\|_{\LPi}^2 + \lambdatwo \|\fhat_m\hnorm{m}^2} + \lambdathree  \|\fhat_m\hnorm{m}^2 ) 
+\sum_{m \in I_0} \lambdathree \|\fhat_m - \fstar_m \hnorm{m}^2 
\notag \\
\leq 
&\max(\phi_s \sqrt{n} \xi_n^2,r) \left( \sum_{m=1}^M \sqrt{\|\fhat_m - \fstar_m \|_{\LPi}^2 + \lambdatmp \|\fhat_m - \fstar_m \hnorm{m}^2 } \right)^2 + 
\frac{1}{n}\sum_{n=1}^n \sum_{m=1}^M\epsilon_i (\fhat_m(x_i) - \fstar_m(x_i))  \notag \\ 
&+  \sum_{m\in I_0} \left(
\frac{3}{2} \lambdaone \sqrt{\|\fhat_m - \fstar_m\|_{\LPi}^2 + \lambdatwo \|\fhat_m - \fstar_m\hnorm{m}^2} + \lambdathree  2 \langle \fstar_m,  \fstar_m - \fhat_m \rangle_{m}
\right).
\label{eq:basicineqLast}
\end{align}
Moreover notice that the assumption \eqref{eq:ForLargen} implies
$\max(\phi_s \sqrt{n} \xi_n^2,r) \leq \frac{1}{2}$. % $2\phi_s \lambdatmp^{\frac{1+q-s}{2}} \leq \frac{1}{2}$. 
Thus \Eqref{eq:ResidDominatedBound} in Lemma \ref{lemm:basicTheoremNormBound} holds.
%implies the following inequality holds 
%\begin{align}
%& \sum_{m=1}^M  \sqrt{\|\fstar_m - \fhat_m\|_{\LPi}^2 + \lambdatwo \|\fstar_m - \fhat_m\hnorm{m}^2}% + \sum_{m\in I_0} \lambdathree \|\fhat_m - \fstar_m\hnorm{m}^2
%\leq
%C_4 \sum_{m\in I_0} \left( \sqrt{\|\fstar_m - \fhat_m\|_n^2 + \lambdatwo \|\fstar_m  - \fhat_m \hnorm{m}^2}\right),
%\label{eq:basicTheoremNormBound2}
%\end{align}
%with probability $1 - \exp( - \sqrt{n})$.

~\\
\noindent{\it Step 1.}
(Bound of the first term in the RHS of \Eqref{eq:basicineqLast}) 
By \Eqref{eq:ResidDominatedBound} in Lemma \ref{lemm:basicTheoremNormBound}, the first term on the RHS of \Eqref{eq:basicineqLast} can be upper bounded as 
\begin{align}
& \max(\phi_s \sqrt{n} \xi_n^2,r) \left( \sum_{m=1}^M \sqrt{\|\fhat_m - \fstar_m \|_{\LPi}^2 + \lambdatmp \|\fhat_m - \fstar_m \hnorm{m}^2 } \right)^2   \notag \\
\leq & \max(\phi_s \sqrt{n} \xi_n^2,r) \left( 8 \sum_{m \in I_0 } \left(1+ \frac{\lambdathree^{\frac{1+q}{2}}\|\gstar_m\hnorm{m}}{\lambdaone}\right) \sqrt{\|\fhat_m - \fstar_m \|_{\LPi}^2 + \lambdatwo \|\fhat_m - \fstar_m \hnorm{m}^2 }  \right)^2  \notag \\
\leq &  128 \max(\phi_s \sqrt{n} \xi_n^2,r) \left( d + \frac{\lambdathree^{1+q} \sum_{m=1}^M \|\gstar_m\hnorm{m}^2}{\lambdaone^2}  \right) \sum_{m \in I_0 } \left( \|\fhat_m - \fstar_m \|_{\LPi}^2 + \lambdatwo \|\fhat_m - \fstar_m \hnorm{m}^2  \right) \notag \\
\leq & 
128 \max(\phi_s \sqrt{n} \xi_n^2,r) \left( d + \frac{\lambdathree^{1+q}\sum_{m=1}^M\|\gstar_m\hnorm{m}^2}{\lambdaone^2}  \right)  \left(\frac{\|\fhat - \fstar \|_{\LPi}^2}{(1- \rho(I_0)^2) \kmin(I_0)} + \sum_{m \in I_0} \lambdatwo \|\fhat_m - \fstar_m \hnorm{m}^2  \right).
\label{eq:firsttermbound}
\end{align}
By assumption, we have $128 \frac{\phi_s \max(\phi_s \sqrt{n} \xi_n^2,r)}{(1- \rho(I_0)^2) \kmin(I_0)} \left( d + \frac{\lambdathree^{1+q}\sum_{m=1}^M\|\gstar_m\hnorm{m}^2}{\lambdaone^2}  \right) \leq \frac{1}{8}$.
Hence the RHS of the above inequality is bounded by 
%\begin{align}
$
\frac{1}{8}   \left(\|\fhat - \fstar \|_{\LPi}^2 + \sum_{m \in I_0} \lambdatwo \|\fhat_m - \fstar_m \hnorm{m}^2  \right).
$
%\label{eq:firsttermbound2}
%\end{align}

~\\
\noindent{\it Step 2.} (Bound of the second term in the RHS of \Eqref{eq:basicineqLast}) 
By \Eqref{eq:ResidDominatedBound} in Lemma \ref{lemm:basicTheoremNormBound}, we have on the event $\scrE_1$
\begin{align}
&\phantom{\leq} \frac{1}{n}\sum_{n=1}^n \sum_{m=1}^M\epsilon_i (\fhat_m(x_i) - \fstar_m(x_i))   \leq \sum_{m=1}^M \eta(t) \phi_s \xi_n \sqrt{\|\fhat_m - \fstar_m \|_{\LPi}^2 + \lambda \|\fhat_m - \fstar_m \hnorm{m}^2}  \notag \\
& \leq \sum_{m\in I_0} 8 \left( 1+ \frac{\lambdathree^{\frac{1+q}{2}}\|\gstar_m\hnorm{m}}{\lambdaone} \right) \eta(t) \phi_s \xi_n \sqrt{\|\fhat_m - \fstar_m \|_{\LPi}^2 + \lambdatwo \|\fhat_m - \fstar_m \hnorm{m}^2  }  \notag \\
& \leq \frac{ 256 \phi_s^2 \eta(t)^2 \xi_n^2  }{ (1- \rho(I_0)^2) \kmin(I_0)}\left( d + \frac{\lambdathree^{1+q}}{\lambdaone^2} \sum_{m=1}^M \|\gstar_m\hnorm{m}^2 \right) \\
&~~~~+ \frac{(1- \rho(I_0)^2) \kmin(I_0)}{8} \sum_{m\in I_0} \left( \|\fhat_m - \fstar_m \|_{\LPi}^2 + \lambdatwo \|\fhat_m - \fstar_m \hnorm{m}^2 \! \! \right)  \notag \\
& \leq \frac{ 16 }{ (1- \rho(I_0)^2) \kmin(I_0)} \left( d \lambdaone^2 \!\! +\! \lambdathree^{1+q} \sum_{m=1}^M \|\gstar_m\hnorm{m}^2 \right) + 
\! \frac{1}{8}\! \left( \|\fhat - \fstar \|_{\LPi}^2 
\! \!+ \! \sum_{m\in I_0} \!\! \lambdatwo \|\fhat_m - \fstar_m \hnorm{m}^2  \right).  \notag \\
\label{eq:secondtermbound}
\end{align}

~\\
\noindent{\it Step 3.} (Bound of the third term in the RHS of \Eqref{eq:basicineqLast})
By Cauchy-Schwarz inequality, we have 
\begin{align}
& \sum_{m\in I_0} \frac{3}{2} \lambdaone \sqrt{\|\fhat_m - \fstar_m\|_{\LPi}^2 + \lambdatwo \|\fhat_m - \fstar_m\hnorm{m}^2} \\
\leq & \frac{9}{2(1- \rho(I_0)^2) \kmin(I_0) } d \lambdaone^2 + 
\frac{(1- \rho(I_0)^2) \kmin(I_0)}{8}\sum_{m\in I_0} \left( \|\fhat_m - \fstar_m \|_{\LPi}^2 +  \lambdatwo \|\fhat_m - \fstar_m \hnorm{m}^2  \right) \notag \\
\leq & \frac{9}{2(1- \rho(I_0)^2) \kmin(I_0) } d \lambdaone^2 + 
\frac{1}{8} \left( \|\fhat - \fstar \|_{\LPi}^2 +  \sum_{m\in I_0} \lambdatwo \|\fhat_m - \fstar_m \hnorm{m}^2  \right).
\label{eq:thirdtermbound}
\end{align}

~\\
\noindent{\it Step 4.} (Bound of the last term in the RHS of \Eqref{eq:basicineqLast})
By \Eqref{eq:gstarYoung}, we have 
\begin{align}
&\sum_{m\in I_0} 2 \lambdathree 
\langle \fstar_m, \fstar_m - \fhat_m\rangle_{\calH_m} %\notag \\
\leq
2 \sum_{m\in I_0}
\lambdathree^{\frac{1+q}{2}} \|\gstar_m \hnorm{m} \sqrt{ \| \fhat_m - \fstar_m \|_{\LPi}^2 + \lambdathree  \| \fhat_m - \fstar_m \hnorm{m}^2} \notag \\
\leq 
& \frac{8  \sum_{m=1}^M \|\gstar_m \hnorm{m}^2}{ (1- \rho(I_0)^2) \kmin(I_0)} d \lambdathree^{1+q}+
 \frac{(1- \rho(I_0)^2) \kmin(I_0)}{8} \sum_{m \in I_0}\left( \| \fhat_m - \fstar_m \|_{\LPi}^2 + \lambdathree  \| \fhat_m - \fstar_m \hnorm{m}^2\right) \notag \\
 \leq 
& \frac{8 \sum_{m=1}^M \|\gstar_m \hnorm{m}^2 }{ (1- \rho(I_0)^2) \kmin(I_0)} d \lambdathree^{1+q}+
 \frac{1}{8} \left( \| \fhat - \fstar \|_{\LPi}^2 + \sum_{m \in I_0} \lambdathree  \| \fhat_m - \fstar_m \hnorm{m}^2\right).
\label{eq:fourthtermbound}
\end{align}

~\\
\noindent{\it Step 5.} (Combining all the bounds)
Substituting the inequalities 
\eqref{eq:firsttermbound}, \eqref{eq:secondtermbound}, \eqref{eq:thirdtermbound} and \eqref{eq:fourthtermbound}
to \Eqref{eq:basicineqLast}, we obtain
\begin{align*}
& \frac{1}{2} \|\fhat - \fstar\|_{\LPi}^2 \\
\leq &  
\frac{ 16}{ (1- \rho(I_0)^2) \kmin(I_0)} \left( d \lambdaone^2 + \lambdathree^{1+q} R_{\gstar}^2\right) 
+ 
 \frac{9}{2(1- \rho(I_0)^2) \kmin(I_0) } d \lambdaone^2  
 +
 \frac{8  R_{\gstar}^2}{ (1- \rho(I_0)^2) \kmin(I_0)} \lambdathree^{1+q} \\
\leq & 
\frac{24}{(1- \rho(I_0)^2) \kmin(I_0) } \left( d \lambdaone^2 +  \lambdathree^{1+q}  \sum_{m=1}^M \|\gstar_m \hnorm{m}^2 \right).
\end{align*}
%where we defined $\widehat{C}_1 = 32 \phi_s^2 \frac{2 C_4^2 + 9/2 + 4 R^2}{(1- \rho(I_0)^2) \kmin(I_0) } $.
This gives the assertion. % by setting $\widehat{C}_1 =  \frac{2 C_4^2 + 9/2 +4}{(1- \rho(I_0)^2) \kmin(I_0) }$.
\end{proof}

\section{Proof of Theorem \ref{eq:LowerboundOfElastMKL}}
\label{proof:LowerboundOfElastMKL}
\begin{proof} {\bf (Theorem \ref{eq:LowerboundOfElastMKL})}
The $\delta$-packing number $\calM(\delta,\calG,L_2(P))$ of a function class $\calG$ with respect to $L_2(P)$ norm is the largest number of functions $\{f_1, \dots, f_{\calM} \} \subseteq \calG$
such that $\|f_i - f_j\|_{L_2(P)} \geq \delta$ for all $i\neq j$.
It is easily checked that 
\begin{equation}
\calN(\delta/2,\calG,L_2(P)) \leq \calM(\delta,\calG,L_2(P)) \leq \calN(\delta,\calG,L_2(P)).
\label{eq:NMrelation}
\end{equation}

First we give the assertion about the $\ell_\infty$-mixed-norm ball (\Eqref{eq:minimaxLinf}).
To simplify the notation, set $R = R_{\infty}$.
For a given $\delta_n > 0$ and $\varepsilon_n > 0$, 
let $Q$ be the $\delta_n$ packing number $\calM(\delta_n,\calH_{\ell_{\infty}}^{d,q}(R),\LPi)$ of $\calH_{\ell_{\infty}}^{d,q}(R)$ and 
$N$ be the $\varepsilon_n$ covering number $\calN(\varepsilon_n,\calH_{\ell_{\infty}}^{d,q}(R),\LPi)$ of $\calH_{\ell_{\infty}}^{d,q}(R)$.
\cite{arXiv:Raskutti+Martin:2010} utilized the techniques developed by \cite{AS:Yang+Barron:99} to show the following inequality in their proof of Theorem 2(b) : 
\begin{align*}
\inf_{\fhat} \sup_{\fstar \in \calH_{\ell_{\infty}}^{d,q}(R)}\EE[\|\fhat - \fstar \|_{\LPi}^2] 
& \geq 
\inf_{\fhat} \sup_{\fstar \in \calH_{\ell_{\infty}}^{d,q}(R)} \frac{\delta_n^2}{2} P[\|\fhat - \fstar \|_{\LPi}^2 \geq \delta_n^2/2] \\
& \geq \frac{\delta_n^2}{2}\left(1-\frac{\log(N) + \frac{n}{2\sigma^2}\varepsilon_n^2 + \log(2)}{\log(Q)}\right).
\end{align*}
Now let %$\tilde{\calH}(\bar{R}):=\{f \in \tilde{\calH} \mid \|f\|_{\tilde{\calH}} \leq \bar{R} \}$ for $\bar{R} >0$ and 
$\tilde{Q}_m := \calM\left(\delta_n/\sqrt{d},\calH_m^q\left(R\right),\LPi\right)$ 
(remind the definition of $\calH_m^q\left(R\right)$ (\Eqref{eq:defHmqR}), and since now $\calH_m$ is taken as $\tilde{\calH}$ for all $m$,
the value $\tilde{Q}_m$ is common for all $m$).
Thus by taking $\delta_n$ and $\varepsilon_n$ to satisfy 
\begin{align}
\frac{n}{2\sigma^2}\varepsilon_n^2 &\leq \log(N), \label{eq:epsilonnbound}  \\
4 \log(N) &\leq \log(Q),   \label{eq:NQbound}
\end{align}
the minimax rate is lower bounded by $\frac{\delta_n^2}{4}$.
In Lemma 5 of \cite{arXiv:Raskutti+Martin:2010}, it is shown that if $\tilde{Q}_1 \geq 2$ and $d \leq M/4$, we have 
$$
\log(Q) \sim d \log(\tilde{Q}_1) + d\log\left(\frac{M}{d}\right).
$$
%and simultaneously we have 
%$$
%\log(N) \sim d \log(\tilde{Q}_1) + d\log\left(\frac{M}{d}\right),
%$$
By the estimation of the covering number of $\calH_m^q(1)$ (\Eqref{eq:coveringconditionHq}), the strong spectrum assumption (\Eqref{eq:strongSpecAss}) and the relation \eqref{eq:NMrelation},
we have 
$$
\log(\tilde{Q}_1) \sim \left(\frac{\delta_n}{R\sqrt{d}}\right)^{-2 \frac{s}{1+q}} = \left(\frac{\delta_n}{R\sqrt{d}}\right)^{-2 \tilde{s}}.
$$
Thus the conditions \eqref{eq:NQbound} and \eqref{eq:epsilonnbound} are satisfied 
if we set $\delta_n = C \varepsilon_n$ with an appropriately chosen constant $C$ and we take $\varepsilon_n$ so that the following inequality holds: 
$$
n \varepsilon_n^2  \lesssim d^{1+\tilde{s}} R^{2\tilde{s}} \varepsilon_n^{-2\tilde{s}} + d\log\left(\frac{M}{d}\right).
$$
It suffices to take 
\begin{align}
\label{eq:epsboundfinal}
\varepsilon_n^2  \sim d  n^{-\frac{1}{1+\tilde{s}}} R^{\frac{2\tilde{s}}{1+\tilde{s}}} + \frac{d\log\left(\frac{M}{d}\right)}{n}.
\end{align}
Note that we have taken $R \geq \sqrt{\frac{\log(M/d)}{n}}$, thus $\tilde{Q}_m \geq 2$ is satisfied if we take the constant in \Eqref{eq:epsboundfinal} appropriately.
Thus we obtain the assertion \eqref{eq:minimaxLinf}.

Next we give the assertion about the $\ell_2$-mixed-norm ball (\Eqref{eq:minimaxL2}).
To simplify the notation, set $R = R_{2}$.
Since $\calH_{\ell_{2}}^{d,q}(R) \supseteq \calH_{\ell_{\infty}}^{d,q}(R/\sqrt{d})$, we obtain 
\begin{align*}
\inf_{\fhat} \sup_{\fstar \in \calH_{\ell_{2}}^{d,q}(R)}\EE[\|\fhat - \fstar \|_{\LPi}^2] 
& \geq 
\inf_{\fhat} \sup_{\fstar \in \calH_{\ell_{\infty}}^{d,q}(R/\sqrt{d})}\EE[\|\fhat - \fstar \|_{\LPi}^2].
\end{align*}
Here notice that we have $\frac{R}{\sqrt{d}} \geq \sqrt{\frac{\log(M/d)}{n}}$ by assumption. 
Thus we can apply the assertion about the $\ell_\infty$-mixed-norm ball \eqref{eq:minimaxLinf} 
to bound the RHS of the just above display. 
We have shown that 
\begin{align*}
\inf_{\fhat} \sup_{\fstar \in \calH_{\ell_{\infty}}^{d,q}(R/\sqrt{d})}\EE[\|\fhat - \fstar \|_{\LPi}^2]
&\gtrsim  d  n^{-\frac{1}{1+\tilde{s}}}  (R/\sqrt{d})^{\frac{2\tilde{s}}{1+\tilde{s}}} + \frac{d\log\left(\frac{M}{d}\right)}{n} \\
&=  d^{\frac{1}{1+\tilde{s}}}  n^{-\frac{1}{1+\tilde{s}}}  R^{\frac{2\tilde{s}}{1+\tilde{s}}} + \frac{d\log\left(\frac{M}{d}\right)}{n}.
\end{align*}
This gives the assertion  \eqref{eq:minimaxL2}.
\end{proof}

%\bibliographystyle{mlapa}
%\bibliography{main,dalmkl}
{
\bibliography{main} %short2}
}

\end{document}